\documentclass{article}

\usepackage{microtype}
\usepackage{graphicx}
\usepackage{booktabs} 

\usepackage{hyperref}

\usepackage[accepted]{icml2024}
\usepackage[utf8]{inputenc} 
\usepackage[T1]{fontenc}    
\usepackage{hyperref}       
\usepackage{url}            
\usepackage{booktabs}       
\usepackage{amsfonts}       
\usepackage{nicefrac}       
\usepackage{microtype}      
\usepackage{xcolor}         
\usepackage{amsmath}
\usepackage{diagbox}
\usepackage{graphics}
\usepackage{epsfig}
\usepackage{threeparttable}
\usepackage{xspace}
\usepackage{siunitx}
\usepackage{graphicx}
\usepackage{amssymb}
\usepackage{threeparttable}
\usepackage{wrapfig}
\usepackage{color}
\usepackage[normalem]{ulem}
\usepackage{multirow}
\usepackage{float}
\usepackage{amsfonts}
\usepackage{bm}
\usepackage{array}
\usepackage{colortbl}
\usepackage{tikz}
\usepackage{diagbox}
\usepackage{rotating}
\usepackage{booktabs}
\usepackage{overpic}
\usepackage{enumitem}
\usepackage{colortbl}
\usepackage[export]{adjustbox}
\usepackage{listings}
\usepackage{pifont}
\usepackage{makecell}
\usepackage{arydshln}

\newcommand{\ie}{\textit{i}.\textit{e}.}
\newcommand{\eg}{\textit{e}.\textit{g}.}
\newcommand{\etc}{\textit{etc}}
\usepackage{booktabs} 
\usepackage{algorithm}
\usepackage{amsmath}
\usepackage{amssymb}
\usepackage{mathtools}
\usepackage{amsthm}
\definecolor{mygray}{gray}{.92}
\DeclareMathOperator*{\argmax}{arg\,max}

\usepackage{makecell}
\usepackage{arydshln}
\usepackage[export]{adjustbox}
\usepackage{listings}
\usepackage{pifont}
\usepackage{diagbox}
\usepackage{graphicx}
\usepackage{amssymb}
\usepackage{threeparttable}
\usepackage{wrapfig}
\usepackage{xspace}
\usepackage{diagbox}
\usepackage{graphics}
\usepackage{epsfig}
\usepackage{threeparttable}
\usepackage{siunitx}
\usepackage{color}
\usepackage[normalem]{ulem}
\usepackage{multirow}
\usepackage{float}
\usepackage{amsfonts}
\usepackage{bm}
\usepackage{array}
\usepackage{rotating}
\usepackage{booktabs}
\usepackage{overpic}
\usepackage{enumitem}
\usepackage{colortbl}
\usepackage{subcaption}
\definecolor{mygray}{gray}{.9}

\makeatletter
\newcommand{\thickhline}{%
	\noalign {\ifnum 0=`}\fi \hrule height 1pt
	\futurelet \reserved@a \@xhline
}
\makeatother

\usepackage[capitalize,noabbrev]{cleveref}
\usepackage{multirow}
\theoremstyle{plain}

\theoremstyle{definition}

\theoremstyle{remark}

\definecolor{mygray}{gray}{.9}
\definecolor{ggray}{RGB}{127,127,127}
\definecolor{reda}{RGB}{192,0,0}
\definecolor{redb}{RGB}{217,148,143}
\definecolor{myyellow}{RGB}{190,144,0}
\definecolor{mygreen}{RGB}{80,100,40}
\definecolor{myblue}{RGB}{30,90,100}
\definecolor{codegreen}{RGB}{79,126,127}
\definecolor{codedefine}{RGB}{153,54,159}
\definecolor{codefunc}{RGB}{73,122,234}
\definecolor{codecall}{RGB}{73,122,234}
\definecolor{codepro}{RGB}{212,96,80}
\definecolor{codedim}{RGB}{89,152,195}
\usepackage[textsize=tiny]{todonotes}

\newtheorem{innercustomthm}{\textbf{Proposition}}
\newenvironment{customthm}[1]
  {\renewcommand\theinnercustomthm{\textbf{\textit{#1}}}\innercustomthm}
  {\endinnercustomthm}

\newtheorem{innercustomthmFormal}{\textbf{Theorem}}
\newenvironment{customthmFormal}[1]
  {\renewcommand\theinnercustomthmFormal{\textbf{\textit{#1}}}\innercustomthmFormal}
  {\endinnercustomthm}

\begin{document}

\twocolumn[
\icmltitle{Prototypical Transformer as Unified Motion Learners}



\icmlsetsymbol{equal}{*}

\begin{icmlauthorlist}
\icmlauthor{Cheng Han}{equal,UMKC,rit}
\icmlauthor{Yawen Lu}{equal,purdue}
\icmlauthor{Guohao Sun}{rit}
\icmlauthor{James C. Liang}{rit}
\icmlauthor{Zhiwen Cao}{purdue}
\icmlauthor{Qifan Wang}{meta}
\icmlauthor{Qiang Guan}{KSU}
\icmlauthor{Sohail A. Dianat}{rit}
\icmlauthor{Raghuveer M. Rao}{DEVCOM}
\icmlauthor{Tong Geng}{UR}
\icmlauthor{Zhiqiang Tao}{rit}
\icmlauthor{Dongfang Liu}{rit}
\end{icmlauthorlist}

\icmlaffiliation{rit}{Rochester Institute of Technology}
\icmlaffiliation{UMKC}{University of Missouri -- Kansas City}
\icmlaffiliation{purdue}{Purdue University}
\icmlaffiliation{meta}{META AI}
\icmlaffiliation{KSU}{Kent State University}
\icmlaffiliation{DEVCOM}{DEVCOM Army Research Laboratory}
\icmlaffiliation{UR}{University of Rochester}

\icmlcorrespondingauthor{Zhiqiang Tao}{zxtics@rit.edu}
\icmlcorrespondingauthor{Dongfang Liu}{dongfang.liu@rit.edu}

\icmlkeywords{Machine Learning, ICML}

\vskip 0.3in
]
\printAffiliationsAndNotice{\icmlEqualContribution} 

\begin{abstract}
In this work,
we introduce the Prototypical Transformer (ProtoFormer), a general and unified framework that approaches various motion tasks from a prototype perspective.
ProtoFormer seamlessly integrates prototype learning with Transformer by thoughtfully considering motion dynamics, introducing two innovative designs.
First, \textit{Cross-Attention Prototyping} discovers prototypes based on signature motion patterns, providing transparency in understanding motion scenes. Second, \textit{Latent Synchronization} guides feature representation learning via prototypes, effectively mitigating the problem of motion uncertainty. Empirical results demonstrate that our approach achieves competitive performance on popular motion tasks such as optical flow and scene depth. Furthermore, it exhibits generality across various downstream tasks, including object tracking and video stabilization. Our code is available
\href{https://github.com/Alvin0629/ProtoFormer}{here}.
\end{abstract}

\section{Introduction}
\label{submission}
\begin{verse}
    \emph{``All is flux, nothing is stationary.''}

    \hfill $-$ Heraclitus~\cite{plato402cratylus}

\end{verse}
The aphorism attributed to Heraclitus underscores the foundation of physics in the natural world. The quest to understand motion holds the potential to unveil the intricate secrets of intelligence~\cite{hawkins2004intelligence}, shedding light on the systematic construction of artificial entities. However, the proliferation of excessively granular motion tasks has catalyzed a fervor for \textit{specialized} models in the realm of deep learning, standing in stark contrast to the enduring scientific tradition — a \textit{generic} solution to elegantly describe physical phenomena in the universe. The following question naturally arises:
\ding{172} Can we discover a \textit{unified} model that serves as a comprehensive motion learner?

Motion learning tasks essentially encompass pixel-level dynamics and correspondence ($e.g.$, optical flow and depth scene estimation). A prevalent challenge in these tasks is the presence of photometric and geometric inconsistencies ($e.g.$, shadow and occlusion), which introduce significant uncertainty during the matching process~\cite{xiong2021self, zhao2020towards}. Consequently, the accuracy of pixel-wise feature matching is compromised, detrimentally impacting the learning of the underlying motion representation.
To address this challenge, a promising solution lies in prototype learning~\cite{smith2002distinguishing, jiang2012recognizing}, where motion measurements are categorized into discrete exemplars. In each exemplar, a prototype functions as a central archetype, capturing the essential attributes of its associated motion patterns observed in the data. The clustering of similar patterns around prototypes can effectively minimize the impact of noise and outlier pixels in feature matching, thereby significantly mitigating the issue of uncertainty. In this context, we can approach question~\ding{172} by exploring:
\ding{173} How can we design a model that incorporates the principles of prototype learning in motion tasks?

\begin{figure}[t]
    \centering
    \includegraphics[width=0.48\textwidth]{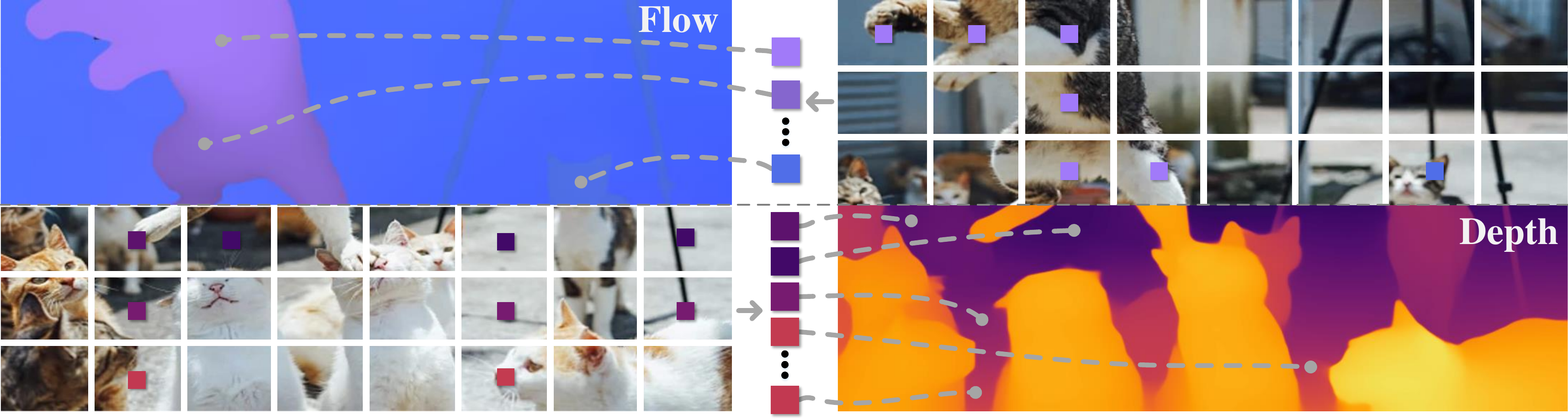}
    \vspace{-7mm}
    \caption{\small{
    \textbf{ProtoFormer as a unified framework}
    considers motion as different levels of dynamics granularity (\eg, instance-driven flow, pixel-anchored depth, \etc).
    \protect\includegraphics[scale=0.20,valign=c]{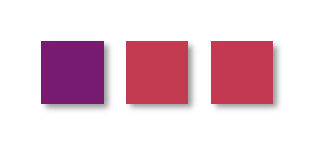} are prototypes.
    }}
    \label{fig:unified}
    \vspace{-7mm}
\end{figure}

Recently, the Transformer architecture has attained ubiquitous adoption, enjoying unambiguous acclaim in both the domains of vision and language~\cite{zhu2021long, yang2023tvt, kim2021vilt}. Its accomplishments are underpinned by the attention mechanism, endowing models with the capability to selectively attend to salient entities within input data.
The capacity to generate context-aware feature representations represents a substantial enhancement of the model’s effectiveness, enabling it to apply as a general solution to
diverse vision tasks. Inspired by its encouraging success, our inquiry naturally delves into a more specific dimension: \ding{174}
How to incorporate the prototype learning capacity into the architecture of Transformer?

To this demand, we employ \underline{Proto}typical Trans\underline{Former} (ProtoFormer) as a unified solution on various motion tasks. Specifically, ProtoFormer
incorporates prototype learning with Transformer.
The method begins by tokenizing images features into patches, where the features are initialized into distinct prototypes.
These prototypes are then recursively updated via \textit{Cross-Attention Prototyping} (\S\ref{subsec:cross-attention}) to capture representative motion characteristics via clustering.
After assignments and updates,
\textit{Latent Synchronization} (\S\ref{subsec:latent-syn}) builds up prototype-feature association, aiming at
denoising and mitigating motion ambiguity. The refined features are later fed into the decoder for task-specific predictions.

Taking the innovations together, ProtoFormer exhibits several compelling attributes. \ding{182} \textbf{Architectural elegance}:
Leveraging a prototype-guided Transformer architecture,
ProtoFormer can handle heterogeneous motion tasks at different levels of dynamics granularity  within the unified fashion (see Fig.~\ref{fig:unified}).
\ding{183} \textbf{Predictive robustness}:
\textcolor{black}{
Prototype learning can inherently diminish the noisy outliers through its density criterion (\S\ref{sec:related_work}).
Anchored by the recursively refined prototypes, feature learning can be further guided towards more robust representations (\S\ref{subsec:latent-syn}). Consequently, it offers a viable solution to the challenge of motion ambiguity (see Fig.~\ref{fig:flow_visual} and Fig.~\ref{fig:depth_visual}).
}
\ding{184} \textbf{Systemic explainability:} The density-based nature from recursive prototyping
offers intuitive visual demonstration of motion prototypes (see Fig.~\ref{fig:visualize} and Fig.~\ref{fig:visualize_dep} in Appendix), enabling direct interpretation of various dynamic patterns sketched by the system.

We conduct a set of comprehensive experiments to evaluate the effectiveness of our approach. In~\S\ref{subsec:optical_flow}, ProtoFormer presents
compelling results on optical flow.
For example, our approach distinctly outperforms CRAFT, achieving 0.48 and 0.69 on the clean and final pass of Sintel, respectively.
In \S\ref{subsec:scene_depth}, we further show the superior performance on depth scene estimation (\eg, 18.6\% improvement in Sintel compared to AdaBins).
Also, visual evidence in \S\ref{subsec:diag-exp}  demonstrates the systemic explainability, which displays direct prototype-pixel correlations.
Results on various downstream tasks including object tracking (\S\ref{sec:object-tracking}) and video stabilization (\S\ref{sec:Video-Stabilization}) are detailed in the Appendix.  We hope our research could provide foundational insights into related fields.

\vspace{-3pt}
\section{Related Work}\label{sec:related_work}
\vspace{-3pt}

\textbf{Motion Task.}
Motion tasks involve intricate processes, encompassing the identification, modeling, and prediction of motion patterns in objects and scenes. These tasks are foundational to diverse computer vision applications, including vehicle and pedestrian motion detection~\cite{shen2023optical, khalifa2020pedestrian, marathe2021evaluating,liang2022triangulation, xu2022vehicle,cui2024collaborative}, abnormal activity detection~\cite{li2021variational, tudor2017unmasking, zhou2019anomalynet}, and video compression~\cite{gao2022structure, hu2021fvc, lu2019dvc}. In the domain of motion tasks, optical flow~\cite{ranjan2017optical,sun2018pwc, teed2020raft, huang2022flowformer, shi2023flowformer++, lu2024promotion} and depth estimation~\cite{bhat2021adabins, patil2022p3depth}, stand out as particularly representative, significantly influencing downstream tasks like object tracking and video stabilization. Current endeavors predominantly focus on task-specific solutions, resulting in duplicated research efforts and suboptimal hardware utilization. In contrast, ProtoFormer stands as a distinctive exploration, aiming to integrate motion tasks under a unified paradigm. This endeavor conceptually differentiates us from existing arts in the field.

\textbf{Prototype Learning.}
Traditionally, prototype learning in machine learning establishes a metric space where features are distinguished by computing their distances/densities to prototypical representations~\cite{lee2023unsupervised}. Early methods include classical approaches like support vector machines~\cite{cortes1995support}, random forest~\cite{breiman2001random}, logistic regression~\cite{hastie2009elements}, etc. With the advent of deep neural networks (DNNs), prototype-based deep learning models find broad applications in few-shot learning~\cite{dong2018few, wang2019panet, liu2020part, yang2020prototype, li2021adaptive,wang2022learning}, zero-shot learning~\cite{jetley2015prototypical, xu2020attribute},  text classification~\cite{farhangi2022protoformer}, and explainable classifiers~\cite{wang2022visual, zhou2022rethinking, qin2023unified}. In the context, we argue: movements within the same object or proximate regions exhibit noteworthy similarities, forming a collective of prototypes. Integrating prototype learning into the model design facilitates the natural encapsulation of diverse dynamic characteristics, enhancing the model's ability to comprehend motion in various contexts.

Moreover, human vision deploys a sophisticated prototyping ability, skillfully focusing on relevant parts of the visual tableau while filtering out extraneous elements~\cite{newell1972human, rudin2022interpretable, giese2003neural}. This feat is realized through Region-of-Interest (RoI) clustering~\cite{meyer2004intensified, mantini2012data}, disassembling discrete pixel entities into salient conceptual groupings. This hierarchical process synthesizes elementary visual elements, like lines, forms, and hues, into complex abstractions representing objects, vistas, and individuals~\cite{kepes1995language}. In an effort to mimic the human visual system, we conceptualize prototype learning from a clustering perspective, iteratively updating and exploring representative prototypes to capture nuanced motion characteristics.

\begin{figure*}
    \centering
    \vspace{-1.2em}
    \includegraphics[width=\textwidth]{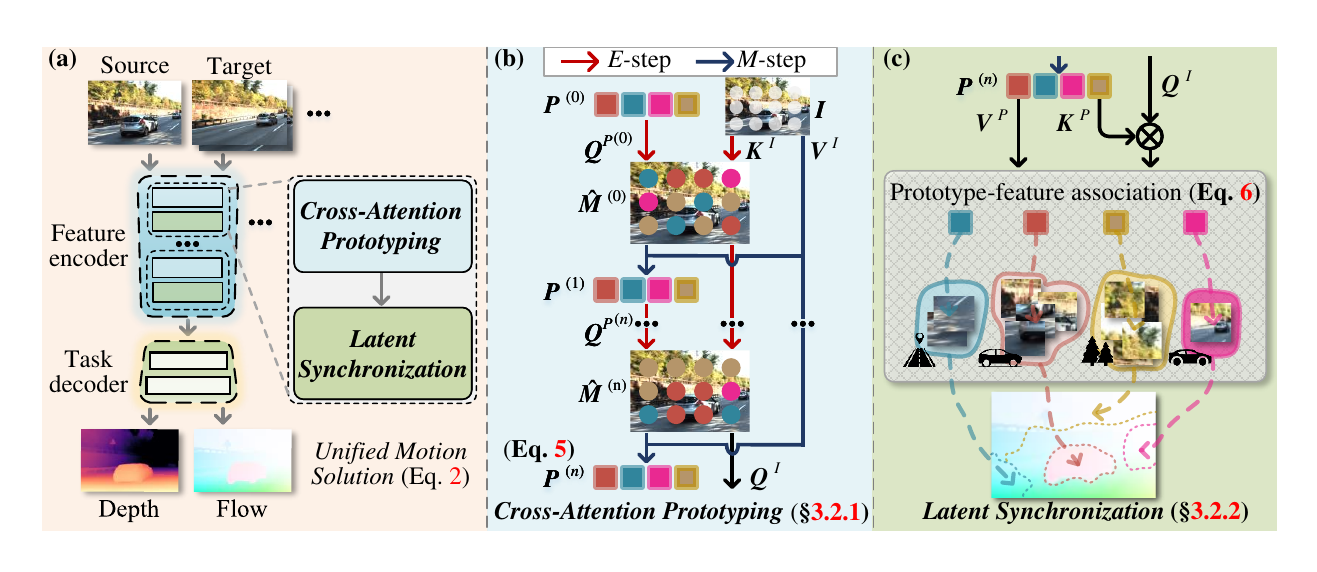}
    \vspace{-1.1cm}
    \caption{\small{(a) \textbf{Overall pipeline of ProtoFormer} (\S\ref{subsec:protoformer}). Movement of a small part of an object within an image is being considered as a rigid motion. In our approach, we use prototypes to understand or predict this kind of motion pattern. (b) In each layer of the \textbf{\textit{Cross-Attention Prototyping}} (see \S\ref{subsec:cross-attention}), there are $N$ sequential iterations encompassing the assignment of feature-prototypes (\ie, $E$-step) and the subsequent updating of these prototypes (\ie, $M$-step) via Eq.~\ref{eq:recurrent}.
    (c) Concurrently, the \textbf{\textit{Latent Synchronization}} process (see \S\ref{subsec:latent-syn}) associates
    the feature representations via the freshly updated motion prototypes,
    (see Eq.~\ref{eq:latent}). For (b) and (c), we apply optical flow for illustration, which demonstrates straightforward systemic explainability. More visualization results are shown in \S\ref{subsec:diag-exp}.}}
    \label{fig:proto}
    \vspace{-4mm}
\end{figure*}

\textbf{Transformer Architecture.}
The transformative impact of Transformers in natural language processing (NLP)~\cite{brown2020language,devlin2018bert,liu2019roberta,raffel2020exploring,vaswani2017attention} has spurred their extensive application in vision-related tasks, including image classification~\cite{dosovitskiy2020image,liu2022swin,liu2021swin,wang2022visual}, and image segmentation~\cite{strudel2021segmenter,wang2021max,wang2021end,zheng2021rethinking}. Transformers excel in visual applications, outperforming convolutional neural networks (CNNs)~\cite{han20232vpt, han2024facing}. This superiority arises from their ability to capture extensive token dependencies in a global context, a limitation of concurrent CNN-based methods that focus on local interactions within convolutional layers~\cite{han2022survey,khan2022transformers,lin2022survey,han20232vpt,cai2022coarse}. The unique attention design in Transformers enables the understanding of global spatial relationships, making them ideal for motion-related tasks where extensive spatial interconnections are pivotal. By amalgamating the attention mechanism with prototype learning, we aim to harness Transformers' representational prowess to unravel intricate patterns in motion tasks, offering a unified, Transformer-based solution.

\vspace{-0.5em}
\section{Methodology}\label{sec:method}
\vspace{-0.5em}

By going through existing literature (\S\ref{sec:related_work}), the integration of prototype learning with the Transformer architecture presents a promising avenue to various motion tasks.
In this section, we first revisit Transformer architecture and reformulate its attention mechanism as prototype learning (\S\ref{subsec:preliminary}).
Based on this insight, we introduce ProtoFormer (\S\ref{subsec:protoformer}),
including two pivotal contributions: \textit{Cross-Attention Prototyping} (\S\ref{subsec:cross-attention}) and \textit{Latent Synchronization} (\S\ref{subsec:latent-syn}), answering question \ding{174}. We elaborate our method below.

\vspace{-0.5em}
\subsection{Preliminary}\label{subsec:preliminary}
\vspace{-2pt}

In our study, we re-conceptualize the Transformer's attention mechanism through the lens of classical clustering; while
the traditional attention map is obtained by computing the
similarity between all query-key pairs~\cite{zhou2021deepvit, han2023flatten}, our approach introduces a density-based  cross-attention estimation, specifically designed to accommodate motion characteristics by aggregating local rigid motion patterns into distinct prototype clusters.

Classic clustering,
a widely embraced paradigm, entails segregating $m$ observations into $k$ distinct groups. It ensures that each observation is aligned with only one cluster that it most closely associates with, based on the highest likelihood or minimal distance (\eg, proximity to the mean).
Formally, clustering can be optimized iteratively between two phases:

\begin{itemize}[leftmargin=*]
\setlength\itemsep{-1.5mm}
\vspace{-0.4cm}
\item \textit{Assignment Phase} allocates each observation to the cluster where it exhibits the maximal probability of belonging or the least spatial separation.
\item \textit{Centroid Recalculation Phase}
recalculates the
centroids of the clusters to reflect the new configuration of the observations within each cluster.
\vspace{-0.4cm}
\end{itemize}

These two phases persist until a point of convergence is reached, indicated either by a cessation in the modification of assignments or by changes that fall beneath a pre-determined threshold, thus implying cluster stabilization.

From a mathematical perspective, assume $\theta$ as the centroids~\cite{wang2022visual} of the clusters, with $x \in \mathcal{X}$ representing an individual observation:
\begin{equation}\small
\vspace{-0.3em}
\theta^{(n+1)} = \argmax_\theta\mathop{\mathbb{E}}(p_{x \sim \mathcal{X}
}(x|\theta^{(n)})).
\label{eq:prelim}
\vspace{-0.3em}
\end{equation}
Here, $\theta^{(n)}$ denotes the $n$-th iteration deduced centroid;
$p(\cdot)$ represents the posterior probability of the data assignments.

\vspace{-2mm}
\subsection{ProtoFormer}\label{subsec:protoformer}
\vspace{-1mm}

The primary objective for ProtoFormer is to optimize the expected likelihood function in the context of clustering within a \textit{unified motion solution} as:
\begin{equation}\small
\hat{\theta}_k = \sum^{K}_{j=1} p(\mathcal{X}|\theta_j) \cdot P(\theta_k,\theta_j),
\label{eq:unified}
\end{equation}
where $\theta$ symbolizes the centroid representations. We recognize the centroid $\theta$ as \textit{Prototype}, which is our optimization target. $K$ represents the total cluster count. The probability $p(\mathcal{X}|\theta_k) \in (0,1)$ are the mixing coefficients for each cluster $k \in \mathcal{K}$, adhering to the constraint $\sum_k p({x|\theta_k}) = 1$. The projected prototype representation $P(\cdot)$ describes the learnable dense vector from a shared parametric family from $k$-th prototype.
This function aggregates across all clusters $\mathcal{K}$, with each projected prototype $P(\theta_k,\theta_j)$
denoting the new projected representations on its respective cluster. $\hat{\theta}_k$ demonstrates the updated prototype considering the prototype representation of $\theta$, and the posterior probability $p(\mathcal{X}|\theta_k)$ (\ie, the conditional likelihood of grouping the data $\mathcal{X}$ given prototype parameterization $\theta_k$).

\subsubsection{Cross-Attention Prototyping via EM clustering}\label{subsec:cross-attention}
To realize prototyping as a unified motion solution on Transformer,
we reformulate the conventional Transformer's self-attention~\cite{vaswani2017attention} into our novel prototypical cross-attention, optimized via $E$xpectation-$M$aximization ($EM$) clustering.
The designed optimization provides a density-based estimation to compute the maximum likelihood for $p_k$ and $\theta_k$,
utilizing posterior probabilities.

\textbf{$E$-Step:} For each observation $x_i \in \mathcal{X}$, $E$-Step computes the $n$-th iteration posterior probabilities $p_k(x_i)$, indicating its affiliation to center $\theta_k$ with the logit vector $s_{x_i,k}$ iteratively as:
\begin{equation}\small
p_k^{(n)}(x_i) = \frac{s_{x_i,k}^{(n)} \cdot P(x_i,\theta_k^{(n)})}{\sum_{j=1}^{K} s_{x_i,j}^{(n)} \cdot P(x_i,\theta_j^{(n)})}.
\end{equation}
$s_{x_i,k}^{(n)}$, $\theta_k^{(n)}$ are the parameters estimated at the $n$-th iteration.

\textbf{$M$-Step:}
Each cluster $\theta_k$ obtains its maximum likelihood estimations $p_k^{(n)}$ and $\theta_k^{(n)}$ from projected sub-sample representations $P'$, updated as:
\begin{equation}\small
\theta_k^{(n+1)} = \frac{1}{N}\sum^{N}_{i=1} \sum^{K}_{j=1} p_k^{(n)}(x_i)\cdot P'(\theta_k^{(n)},\theta_j^{(n)}).
\label{eq:e_step_theta}
\end{equation}
In practice, given feature embeddings $\bm{I}\in\mathbb{R}^{HW \times D}$ and set initial cluster centers $\bm{P}^{(0)}$ as $K$ prototypes,
we encapsulate the discussed $EM$ clustering process within a \textit{Cross-Attention Prototyping} layer (see Fig.~\ref{fig:proto}(b)) with $N$ iterations:
\begin{equation}\label{eq:recurrent}\small
	\begin{aligned}
E\text{-step:~~~~~~} \hat{\bm{M}}^{(n)} &= \mathop{\mathrm{softmax}}\nolimits_K(\bm{Q}^{P^{(n)}}(\bm{K}^I)^\top),\\
M\text{-step:~~~} \bm{P}^{(n+1)} &= \hat{\bm{M}}^{(n)}\bm{V}^I\!\in\!\mathbb{R}^{K \times D},
	\end{aligned}
\end{equation}
where $n \in \{1, \cdots, N\}$. $\hat{\bm{M}} \in [0,1]^{K \times HW}$ is the ``soft'' pixel-prototype assignment matrix, representing probability maps of
prototypes. $\bm{Q}^{P} \in \mathbb{R}^{K \times D}$ is the query vector projected from the prototype representation $\bm{P}$, and $\bm{V}^{I}, \bm{K}^{I} \in \mathbb{R}^{HW \times D}$ are the value and key vectors projected from the image features $\bm{I}$, respectively. Our proposed layer can thus update the prototyping membership $\hat{\bm{M}}$ (\ie, $E$-step) and the prototypes $\bm{P}$ (\ie, $M$-step) iteratively.
The key characteristic of this approach is its assurance of an incremental convergence in the likelihood function with each iteration~(see Eq.~\ref{eq:e_step_theta}).
In essence, the $E$-step evaluates the current membership of the data representations based on existing prototypes, while the $M$-step refines the prototypes to align with pixels, ensuring a steady progression towards optimal clustering. By performing cross-attention prototyping on the source and target images separately, it addresses the complexities associated with motion uncertainty and photometric inconsistency. We also modify the default \textit{softmax} operator from $HW$ to $K$, mimicking the $EM$ clustering.

The proposed layer enjoys several compelling features:
\begin{itemize}[leftmargin=*]
\setlength\itemsep{-1.5mm}
\vspace{-4.0mm}
\item \textit{Convergence:} $EM$ clustering monotonically improves the marginal likelihood and is empirically validated to converge towards a local optimum~\cite{vattani2009k, ikotun2023k, balakrishnan2017statistical}, given a sufficient number of iterations (\textit{Proof} is provided in Appendix \S\ref{sec:proofs}).
\begin{customthm}{1}\label{prop:bound}
Suppose that the EM operator is contractive with parameter $\kappa \in (0,1)$ on the ball $\mathcal{B}_{2}(r;\theta)$, and the initial vector $\theta^{(0)}$ belongs to $\mathcal{B}_{2}(r;\theta)$. For a given iteration $N$, when the sample size $m$ is large enough to ensure $\epsilon_{M}(\frac{m}{N}, \frac{\delta}{N}) \leq (1-\kappa)r$. Then the EM iterates $\left\{\theta^{(n)}\right\}^{N}_{n=0}$ based on $\frac{m}{N}$ samples per round satisfy the bound that:
\begin{equation}
\small
    ||\theta^{(n)} - \hat{\theta}||_{2} \leq \kappa^{n} ||\theta^{(0)} - \hat{\theta}||_{2} + \frac{1}{1-\kappa} \epsilon_{M}(\frac{m}{N}, \frac{\delta}{N}).
\end{equation}
\end{customthm}
In this context, our proposed \textit{Cross-Attention Prototyping} leverages the strengths of recursive clustering, iterated over $N$ steps. This approach significantly enhances the likelihood of converging to an optimal configuration for motion partitioning (see \S\ref{subsec:diag-exp}).
\item \textit{Transparency:} Prototyping emerges as an indispensable mechanism for contextual understanding from motion scenes,
recognizing and grouping similar patterns and movements.
It clusters pixels as prototypes that display homogeneity in characteristics such as flow or depth.
By aggregating entities that exhibit shared attributes, the prototypes are able to describe the intrinsic dynamics.
Furthermore, prototyping
provides a foundational schema for motion comprehension, wherein each prototype embodies a microcosm of the objects within the scene, encapsulating unique elements and their interrelations.

\item \textit{Efficiency:} \textit{Cross-Attention Prototyping} operates with a time complexity of $\mathcal{O}(NKHWD)$, showing a significant improvement over the self-attention mechanism with $\mathcal{O}(H^2W^2D)$ (see \S\ref{subsec:diag-exp}). The foundation lies in the relationship that $NK \ll HW$ (\eg, 60 $vs$ 25920 in the first stage with image of $960 \times 432$ resolution). This difference becomes especially pronounced in pyramid architectures~\cite{wang2021pyramid,liu2021swin, liang2023clustseg, liang2024clusterfomer}, where
the total number of $NK$ tends to be substantially smaller than $HW$, particularly in the early stages of the network.
In each iteration, only the query matrix $\bm{Q}$ requires an update; the key $\bm{K}$ and value $\bm{V}$ matrices are computed just once. This selective updating significantly
reduces the computational load
particularly beneficial in handling large-scale data in high-dimensional feature spaces.

\end{itemize}

\vspace{-2mm}
\subsubsection{Prototype-Feature Corresponding by Latent Synchronization}
\label{subsec:latent-syn}
\vspace{-1mm}
We further refine the
feature representations, synchronizing the projection of $K$ prototypes into a $H \times W$ feature representation. This approach aligns the prototype representations with motion features (see Fig.~\ref{fig:proto}(c)).

The main technique lies in the Feed-Forward Network (FFN) that incorporates with a masked cross-attention mechanism:
\begin{equation}\small
\hat{I} = \text{FFN}(\text{Cross-Attention}(Q^I, K^P, V^P, \mathcal{M}_P)),
\label{eq:latent}
\end{equation}
where $\mathcal{M}_P$ stands for the feature assignment mask maps based on the similarity of corresponding prototypes $P$.
$\hat{I}$ represents the refined feature, $Q^I$ denotes the query projection derived from the input image feature, $K^P$ and $V^P$ represent the key and value projections sourced from the learning prototypes, respectively. \textit{Latent Synchronization} primarily aims at augmenting the feature learning for prototype-feature association,
reducing motion ambiguity.
It facilitates the extraction and encapsulation of each feature representation's latent distribution within its own
prototype.

\textit{Latent Synchronization} enjoys appealing characteristics:
\begin{itemize}[leftmargin=*]
\setlength\itemsep{-1mm}
\item \textit{Blended Paradigm:}
\textcolor{black}{\textit{Latent Synchronization} blends unsupervised prototype mining (\S\ref{subsec:cross-attention}) and supervised feature representation learning (\S\ref{subsec:latent-syn}) in a synergy $-$ local significant motion patterns are automatically explored to facilitate density-based prototyping; the supervisory signal from task-specific heads directly optimizes the representation, which in turn boosts meaningful prototyping.}

\item \textit{Prototype-Anchored Learning:}
Density-based prototype learning computes reliable probabilities in prototype assignment recursively (Eq.~\ref{eq:recurrent}). Anchored by the updated prototypes, the feature is further guided towards more robust representations via prototype-feature association (Eq.~\ref{eq:latent}). Consequently, the motion patterns are more likely to center around areas of high data density, which in turn boosts robustness towards motion ambiguity.
\end{itemize}

\begin{table*}[t]
\begin{adjustbox}{width=0.78\width,center}
\begin{tabular}{c||l|cc|cc|cc|c}
\hline \thickhline
\rowcolor{mygray}
&  & \multicolumn{2}{c|}{Sintel (train)} & \multicolumn{2}{c|}{KITTI-15 (train)} & \multicolumn{2}{c|}{Sintel (test)} & KITTI-15 (test) \\
\cline{3-9}
\rowcolor{mygray}
\multirow{-2}{*}{Training}  & \multirow{-2}{*}{Method} & Clean $\downarrow$ & Final $\downarrow$ & F1-epe $\downarrow$ & F1-all $\downarrow$ & Clean $\downarrow$ & Final $\downarrow$ & F1-all $\downarrow$ \\
 \hline \hline
\multirow{2}{*}{A} & Perceiver IO~\cite{jaegle2021perceiver} & 1.81 & 2.42 & 4.98 & - & - & - & - \\
 & RAFT-A~\cite{RAFT-A}  & 1.95 & 2.57 & 4.23 & - & - & - & - \\
 \hline
\multirow{12}{*}{C+T} & RAFT~\cite{teed2020raft} & 1.43 & 2.71 & 5.04 & 17.4 & - & - & - \\
 & Separable Flow~\cite{Separable}  & 1.30 & 2.59 & 4.60 & 15.9 & - & - & - \\
 & GMA~\cite{jiang2021learning} & 1.30 & 2.74 & 4.69 & 17.1 & - & - & - \\
 & AGFlow~\cite{agflow} & 1.31 & 2.69 & 4.82 & 17.0 & - & - & - \\
 & KPA-Flow~\cite{kpaflow} & 1.28 & 2.68 & 4.46 & 15.9 & - & - & - \\
 & DIP~\cite{dip} & 1.30 & 2.82 & 4.29 & 13.7 & - & - & - \\
 & GMFlowNet~\cite{gmflownet} & 1.14 & 2.71 & 4.24 & 15.4 & - & - & - \\
 & GMFlow~\cite{xu2022gmflow} & 1.08 & 2.48 & 7.77 & 23.4 & - & - & - \\
 & CRAFT~\cite{sui2022craft} & 1.27 & 2.79 & 4.88 & 17.5 & - & - & - \\
 & FlowFormer~\cite{huang2022flowformer} & 1.01 & 2.40 & 4.09 & 14.7 & - & - & - \\
 & SKFlow~\cite{Zhai2019SKFlowOF} & 1.22 & 2.46 & 4.27 & 15.5 & - & - & - \\
 & MatchFlow~\cite{dong2023rethinking} &1.14&2.71&4.19&13.6&-&-&-\\
 & \cellcolor{mygray}\textbf{ProtoFormer (Ours)} & \cellcolor{mygray}1.04 & \cellcolor{mygray}2.43 & \cellcolor{mygray}4.08 & \cellcolor{mygray}14.6 & - & - & - \\
 \hline
\multirow{13}{*}{C+T+S+K+H} & RAFT~\cite{teed2020raft} & 0.76 & 1.22 & 0.63 & 1.5 & 1.61 & 2.86 & 5.10 \\
 & RAFT-A~\cite{RAFT-A} & - & - & - & - & 2.01 & 3.14 & 4.78 \\
 & Separable Flow~\cite{Separable} & 0.69 & 1.10 & 0.69 & 1.60 & 1.50 & 2.67 & 4.64 \\
 & GMA~\cite{jiang2021learning} & 0.62 & 1.06 & 0.57 & 1.2 & 1.39 & 2.47 & 5.15 \\
 & AGFlow~\cite{agflow} & 0.65 & 1.07 & 0.58 & 1.2 & 1.43 & 2.47 & 4.89 \\
 & KPA-Flow~\cite{kpaflow} & 0.60 & 1.02 & 0.52 & 1.1 & 1.35 & 2.36 & 4.60 \\
 & DIP~\cite{dip} & - & - & - & - & 1.44 & 2.83 & 4.21 \\
 & GMFlowNet~\cite{gmflownet} & 0.59 & 0.91 & 0.64 & 1.51 & 1.39 & 2.65 & 4.79 \\
 & GMFlow~\cite{xu2022gmflow} & - & - & - & - & 1.74 & 2.90 & 9.32 \\
 & CRAFT~\cite{sui2022craft} & 0.60 & 1.06 & 0.58 & 1.34 & 1.45 & 2.42 & 4.79 \\
 & Flowformer~\cite{huang2022flowformer} & 0.48 & 0.74 & 0.53 & 1.11 & 1.20 & 2.12 & 4.68 \\
 & SKFlow~\cite{Zhai2019SKFlowOF} & 0.52 & 0.78 & 0.51 & 0.94 & 1.28 & 2.23 & 4.87 \\
 & MatchFlow~\cite{dong2023rethinking} &0.51&0.81&0.59&1.3&1.33&2.64&4.72 \\
 & \cellcolor{mygray}\textbf{ProtoFormer (Ours)} & \cellcolor{mygray}0.48 & \cellcolor{mygray}0.69 & \cellcolor{mygray}0.50 & \cellcolor{mygray}1.09 & \cellcolor{mygray}1.06  & \cellcolor{mygray}2.07  & \cellcolor{mygray}4.35               \\
\hline
\end{tabular}
\end{adjustbox}
\vspace{-2mm}
\caption{\small{\textbf{Quantitative results on standard Sintel and KITTI flow benchmarks.} `A' denotes the Autoflow dataset; `C + T' denotes training on the FlyingChairs and FlyingThings datasets only; `C + T + S + K + H' fine-tunes on a combination of Sintel, KITTI, and HD1K training sets. Error metrics are lower is better with "$\downarrow$", and accuracy metrics are higher is better with "$\uparrow$". Same for Table~\ref{depth_tab}.}}
\label{flow_tab}
\vspace{-5mm}
\end{table*}

\vspace{-1em}
\subsection{Implementation Details}
ProtoFormer is built upon Twins architecture~\cite{chu2021twins}. Detailed training and testing configurations are provided in \S\ref{sec:config}. The key components (see Fig.~\ref{fig:proto}) are:
\begin{itemize}[leftmargin=*]
\setlength\itemsep{-1.5mm}
\vspace{-4mm}
    \item \textit{Feature Encoder} contains two stages with window sizes of 4 and 8, respectively, which convert input images into features. We follow the common practice~\cite{chu2021twins} and utilize two blocks within each stage. In addition, we reformulate the vanilla self-attention into our cross-attention prototyping layers (\S\ref{subsec:cross-attention}) to recursively update the initialized prototypes. Once these prototypes have been updated, the latent synchronization layer (\S\ref{subsec:latent-syn}) augments the feature learning via prototype-feature association, reducing motion ambiguity.

    \item \textit{Task Decoder} is designed for task-specific motion predictions.
    We follow the design~\cite{huang2022flowformer, zhou2021r} for flow and depth.

    \item \textit{Cross-Attention Prototyping} reformulates the vanilla self-attention layers by the $EM$ cross-attention clustering process for prototyping learning (\S\ref{subsec:cross-attention}). Within each Cross-Attention Prototyping layer, twenty prototypes and
    three iterations are conducted as default (\S\ref{subsec:diag-exp}).

    \item \textit{Latent Synchronization} updates the feature map in accordance with the prototypes present in the latent feature space
    (\S\ref{subsec:latent-syn}) by applying Feed-Forward Networks and incorporating a masked cross-attention mechanism.
\end{itemize}

\vspace{-4mm}
\section{Experiments}\label{sec:Experiment}
\vspace{-3pt}
We comprehensively examine the performance and unity of our proposed ProtoFormer on two representative motion tasks, including optical flow (see \S\ref{subsec:optical_flow}) and scene depth (see \S\ref{subsec:scene_depth}).
In our pursuit of a unified solution for motion-related tasks, we not only underscore the merit of such integration but also exhibit superior performance, and to further enhance the \textit{paradigmatic generalization}, we broaden its application to object tracking (\S\ref{sec:object-tracking}) and video stabilization (\S\ref{sec:Video-Stabilization}) in Appendix, reaching \textit{competitive} performance.

\vspace{-3pt}
\subsection{Experiments on Optical Flow}\label{subsec:optical_flow}
\vspace{-3pt}

\noindent \textbf{Datasets.}
Following the previous works~\cite{huang2022flowformer,dong2023rethinking}, we first train the proposed method on FlyingChair~\cite{dosovitskiy2015flownet} and FlyingThings~\cite{mayer2016large}, and then fine-tune it on a large combination of datasets (C+T+S+K+H) to allow evaluation on the Sintel and KITTI-2015 benchmarks.
\\
\noindent \textbf{Metrics.}
To facilitate a fair comparison, we adopt the commonly used metric, the average end-point-error (F1-epe), measuring the average $\mathit{l}_{2}$ distance between the prediction and ground truth, and the percentage of outliers over all pixels (F1-all), which describes the error exceeding 3 pixels or 5\% $w.r.t.$ the ground truth, for optical flow estimation.
\\
\noindent \textbf{Quantitative Results.}
Table~\ref{flow_tab} reports the evaluation results of our model on the Sintel and KITTI datasets. The results under the `C+T' setting reflect the generalization capability of our ProtoFormer, where it achieves 1.04 and 2.43 on the clean and final pass of Sintel, improving the recently popular CRAFT~\cite{sui2022craft} by 0.23 and 0.36.
After training under the mixed setting of `C+T+S+K+H', the proposed prototype-based model achieves 1.06 and 2.07 on the clean and final pass of Sintel and 4.35 F1-epe on KITTI.
\begin{figure*}
    \centering
    \includegraphics[width=0.91\textwidth]{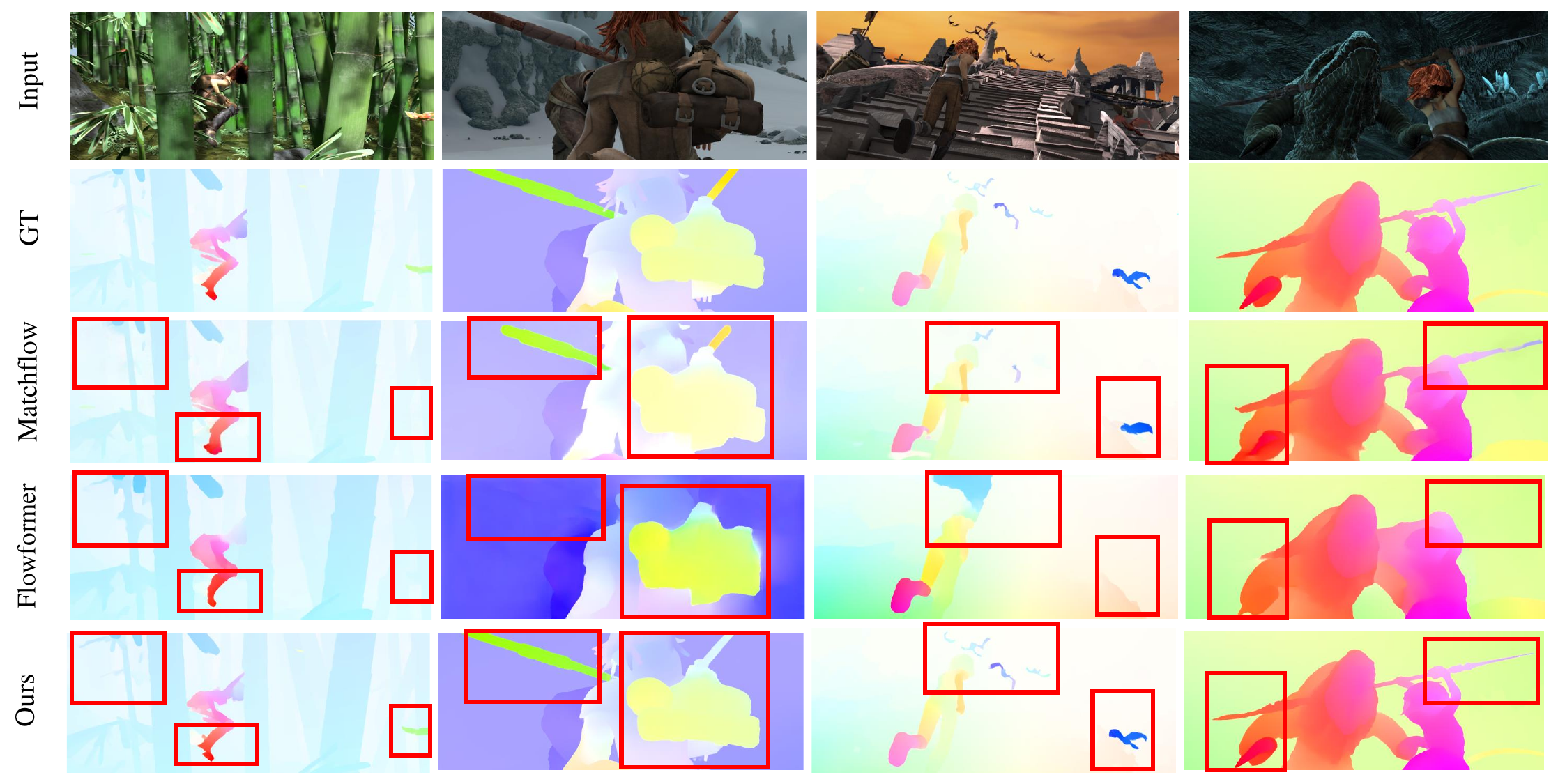}
    \vspace{-4mm}
    \caption{\small{\textbf{Qualitative results on the Sintel.} The red boxes highlight the regions compared. Matchflow~\cite{dong2023rethinking} appears blurry and ambiguous on textureless and occluded objects, while Flowformer~\cite{huang2022flowformer} fails to recover complete and detailed information. Ours can estimate clear and complete flow motion, which is closer to ground truth.
}}
    \label{fig:flow_visual}
    \vspace{-5mm}
\end{figure*}
\\
\noindent \textbf{Qualitative Results.}
Fig.~\ref{fig:flow_visual} shows the qualitative results on the Sintel flow dataset. As seen, ProtoFormer shows more global and finer details on both object and motion boundaries, without being affected by the shadows and textureless surfaces. In the first and fourth examples, our model recovers full shape and fine details remarkably well, \eg, on bamboo and weapons. This is in stark contrast to other methods, which struggle with producing clear predictions, primarily due to challenges posed by occlusions and variations in illumination.
In the second and third examples, we show the significantly consistent estimation of the occluded and textureless regions, \eg, the backpack and birds in the sky. The highlighted regions with red boxes prove the efficacy of ProtoFormer in object clustering and avoiding motion ambiguity. More results are shown in Fig.~\ref{fig:more_flow}.

\vspace{-3pt}
\subsection{Experiments on Scene Depth}\label{subsec:scene_depth}
\vspace{-3pt}

\begin{figure*}
    \centering
    \includegraphics[width=0.92\textwidth]{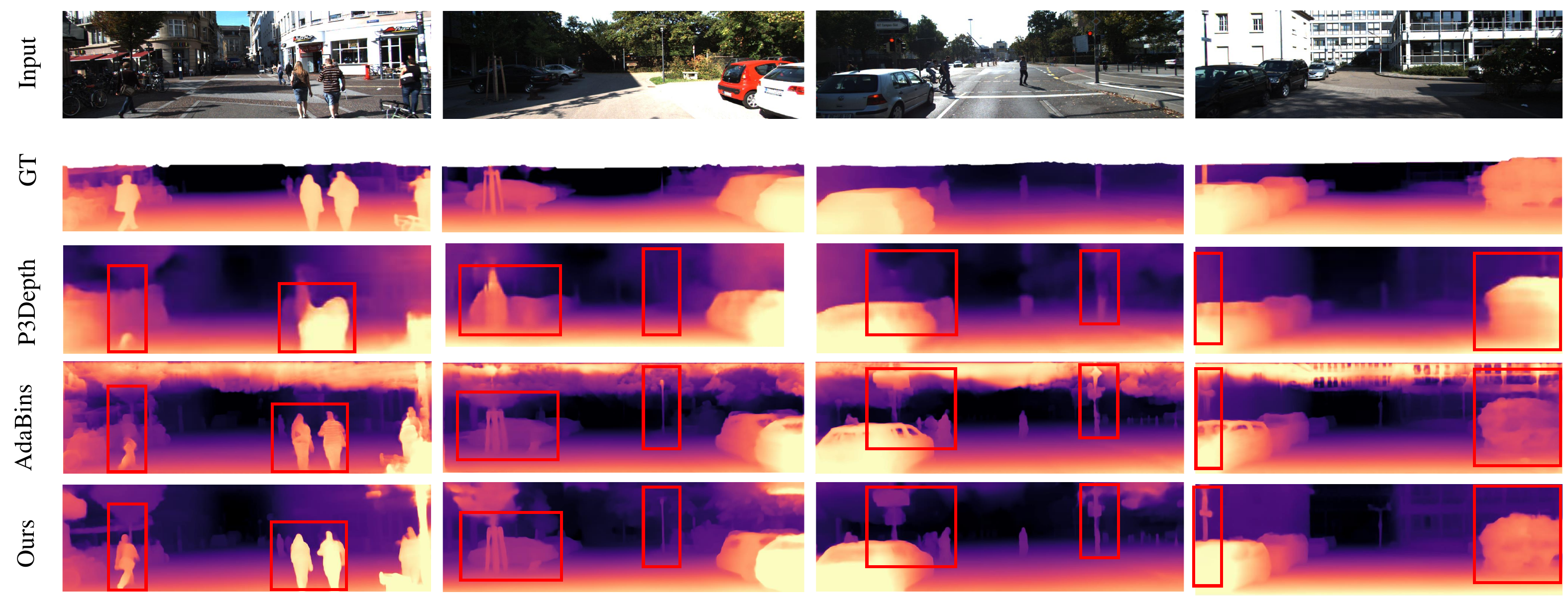}
    \vspace{-3mm}
    \caption{\small{\textbf{Qualitative depth comparison results on the KITTI.} The red boxes indicate the highlighted regions. P3Depth~\cite{patil2022p3depth} and AdaBins~\cite{bhat2021adabins} have limited receptive fields and do not consider conceptual object-level groupings, thus producing discontinuous and ambiguous predictions. While ours can estimate consistent and sharp depths, which is closer to ground truth.}
}
    \label{fig:depth_visual}
\end{figure*}

\noindent \textbf{Datasets.} Similar to optical flow, we evaluate ProtoFormer on both real and synthetic datasets, using KITTI~\cite{geiger2013vision} and MPI Sintel~\cite{Butler2012} for evaluation. As an autonomous driving dataset consisting of 61 outdoor scenes of various modalities, we use the KITTI Eigen depth split, which contains a standard depth estimation split proposed by Eigen et al.~\cite{eigen2014depth} consisting of 32 scenes for training and 29 scenes for testing. The MPI Sintel is a long synthetic stereo sequence with a large motion and depth range and contains a total of 35 rendered sequences.
\\
\noindent \textbf{Metrics.}
We follow the standard metrics of absolute relative error (Abs Rel), root mean square error (RMSE), and the percentage of inlier pixels with $\delta_{1} < \tau \ (\tau = 1.25)$.
\\
\noindent \textbf{Quantitative Results.}
Table~\ref{depth_tab} shows the test results on Sintel and KITTI. Our model achieved the best performance on most of the error and accuracy metrics, proving its powerful feature representation and generalization ability to different scenarios.
Compared with recent directly supervised depth estimation methods~\cite{eigen2014depth,fu2018deep,bhat2021adabins}, ProtoFormer demonstrates significantly superior capability, benefited from the incorporation of prototypical learning and cross-attention architecture. Even \textit{without} using additional constraints and priors such as surface normal and piecewise planarity in~\cite{yin2019enforcing,patil2022p3depth},
our model improves the performance of concurrent P3Depth in error by a large margin on KITTI.
Furthermore, our approach demonstrates superior performance compared to methods that implement self-supervised consistency and strategies~\cite{godard2017unsupervised,zhang2023lite}. This underscores the efficacy of our model, which exhibits exceptional adaptability and learning capacity for fine-tuning across more general motion-related tasks.
\begin{table}[htb]
\resizebox{\columnwidth}{!}{
\setlength\tabcolsep{4pt}
\begin{tabular}{l||ccc|ccc}
\hline \thickhline
\rowcolor{mygray}
\cellcolor{mygray}                         & \multicolumn{3}{c|}{\cellcolor{mygray}Sintel} & \multicolumn{3}{c}{\cellcolor{mygray}KITTI} \\
\rowcolor{mygray}
\multirow{-2}{*}{\cellcolor{mygray}Method} & Abs Rel $\downarrow$     & RMSE $\downarrow$      & Sq Rel $\downarrow$    & Abs Rel $\downarrow$     & RMSE $\downarrow$     & $\delta_{1}$ $\uparrow$   \\ \hline \hline
Eigen et al.                                      & 0.797       & 0.834     & 0.703                     & 0.203       & 6.307    & 0.702                    \\
Godard et al.                                    & -           & -         & -                        & 0.114       & 4.935    & 0.861                    \\
Fu et al.                                        & -           & -         & -                        & 0.072       & 2.727    & 0.932                    \\
Yin et al.                                       & 0.746        & 0.611      & 0.652                     & 0.072       & 3.258    & 0.938                    \\
AdaBins                                          & 0.730       & 0.572     & 0.647                     & 0.067       & 2.960    & 0.949                    \\
P3Depth                                          & 0.653       & 0.396     & 0.571                     & 0.071       & 2.842    & 0.953                    \\
\rowcolor{mygray}
\textbf{Ours}                                    & 0.594        & 0.486      & 0.538                     & 0.062       & 2.716     & 0.949     \\ \hline
\end{tabular}
}
\vspace{-2mm}
\caption{\small{\textbf{Quantitative results on Sintel and KITTI depth datasets.} With both test data unseen by the model, we can achieve leading performance over state-of-the-art methods \cite{eigen2014depth, godard2017unsupervised, fu2018deep, yin2019enforcing, bhat2021adabins, patil2022p3depth}.}}
\label{depth_tab}
\vspace{-2mm}
\end{table}
\\
\noindent \textbf{Qualitative Results.} Fig.~\ref{fig:depth_visual} shows the qualitative depth comparison on the KITTI Eigen depth datasets.
Our ProtoFormer demonstrates superior capability in delineating object surface contours, particularly in scenarios involving dynamic entities such as pedestrians and vehicles, as well as in capturing the finer details of objects like traffic signs and light poles.
For example, for the moving pedestrians and vehicles in Sample 1 and Sample 3, ProtoFormer estimates a more consistent and complete depth on object surfaces and provides a clearer boundary than P3Depth~\cite{patil2022p3depth} and AdaBins~\cite{bhat2021adabins}. For the plant stand and traffic poles in Sample 2, Sample 3 and Sample 4, our methods separates other methods evidently from the noisy and complex scene backgrounds. These demonstrated the superiority of incorporating prototype learning into depth training to perceive geometric consistent and mitigate motion ambiguity.
More visual evidences are provided in Fig.~\ref{fig:more_depth}.

\vspace{-5pt}
\subsection{Diagnostic Experiments}\label{subsec:diag-exp}
\vspace{-3pt}
This section ablates ProtoFormer's key components and configurations. More ablations are included in Appendix \S\ref{sec:scene-depth-ablative}.
\\
\begin{table*}[t]
\vspace{-5pt}
    \caption{\small{A set of \textbf{ablative studies} on optical flow (see \S\ref{subsec:diag-exp}). The best performances are marked in \textbf{bold}.}}
     \vspace{-5pt}
     \begin{subtable}{0.47\linewidth}
    \vspace{-30pt}					
     \captionsetup{width=.95\linewidth}
						\resizebox{\textwidth}{!}{
							\setlength\tabcolsep{4pt}
							\renewcommand\arraystretch{1.1}
							\begin{tabular}{l|c|cc}
								\thickhline
								\rowcolor{mygray}
								Algorithm Component &  \#Params& Sintel clean & Sintel final\\

								\hline\hline
                                \texttt{Base}&9.63M&0.55&0.81\\							\arrayrulecolor{gray}\hdashline\arrayrulecolor{black}	
                                ~~~$+$ Cross-Attention Prototyping &  11.57M & 0.51 & 0.74\\
                                ~~~$+$ Latent Synchronization & 10.26M & 0.53 & 0.77\\
								\hline
        \arrayrulecolor{gray}\arrayrulecolor{black}
        \textbf{ProtoFormer} (\textbf{All included})& 11.90M &0.48&0.69\\
        \hline
						\end{tabular}}
						\setlength{\abovecaptionskip}{0.3cm}
						\setlength{\belowcaptionskip}{-0.1cm}
						\caption{\small{Key Component Analysis}}
						\label{table:key}
					\end{subtable}
          \begin{subtable}{0.49\linewidth}
						\captionsetup{width=.95\linewidth}
						\resizebox{\textwidth}{!}{
							\setlength\tabcolsep{4pt}
							\renewcommand\arraystretch{1.1}
							\begin{tabular}{l|c|cc}
								\thickhline
								\rowcolor{mygray}
								Variant Prototype Updating Strategy&  \#Params& Sintel clean & Sintel final\\

								\hline\hline
                                Cosine Similarity&10.28M&0.51&0.75\\							
                                Vanilla Cross-Attention~\cite{vaswani2017attention}&14.88M& 0.50&0.73\\
                                Criss Cross-Attention~\cite{huang2019ccnet}&14.56M&0.50&0.72\\
                                $K$-Means~\cite{yu2022k}&11.81M&0.49&0.71\\
								\hline
        \arrayrulecolor{gray}\arrayrulecolor{black}
        \textbf{Cross-Attention Prototyping (Eq.~\ref{eq:recurrent})}& 11.90M &0.48&0.69\\
        \hline
						\end{tabular}}
						\setlength{\abovecaptionskip}{0.1cm}
						\setlength{\belowcaptionskip}{-0.1cm}
						\caption{\small{\textit{Cross-Attention Prototyping}}}
						\label{table:recurrent}
					\end{subtable}
     \vspace{-0.8em}
     \begin{subtable}{0.30\linewidth}
					\vspace{-6pt}	\captionsetup{width=.95\linewidth}
						\resizebox{\textwidth}{!}{
							\setlength\tabcolsep{4pt}
							\renewcommand\arraystretch{1.1}
							\begin{tabular}{c|c|cc}
								\thickhline
								\rowcolor{mygray}
								\#Iterations ($N$) &  \#Params& Sintel clean & Sintel final\\

								\hline\hline
                                1&\multirow{4}{*}{11.90M}&0.52&0.75\\							
                                2&&0.49&0.71\\
                                \textbf{3}&  &0.48&0.69\\
                                4&&0.48&0.68\\
        \hline
						\end{tabular}}
						\setlength{\abovecaptionskip}{0.3cm}
						\setlength{\belowcaptionskip}{-0.1cm}
						\caption{\small{Number of Iterations}}
						\label{table:number}
					\end{subtable}
     \begin{subtable}{0.30\linewidth}
     \vspace{-6pt}
			\captionsetup{width=.95\linewidth}
						\resizebox{\textwidth}{!}{
							\setlength\tabcolsep{4pt}
							\renewcommand\arraystretch{1.1}
							\begin{tabular}{c|c|cc}
								\thickhline
								\rowcolor{mygray}
								\#Prototypes ($K$) &  \#Params& Sintel clean & Sintel final\\
								\hline\hline
                                10 & 8.95M & 0.53&0.78\\				
                                50 & 9.78M & 0.51 & 0.73\\
                                \textbf{100}& 11.90M &0.48&0.69\\
                                200 & 14.21M & 0.49 & 0.71 \\
								\hline
						\end{tabular}}
						\setlength{\abovecaptionskip}{0.3cm}
						\setlength{\belowcaptionskip}{-0.1cm}
						\caption{\small{Number of Prototypes}}
						\label{table:prototype-number}
					\end{subtable}
          \begin{subtable}{0.40\linewidth}
						\vspace{-6pt}
      \captionsetup{width=.95\linewidth}
						\resizebox{\textwidth}{!}{
							\setlength\tabcolsep{4pt}
							\renewcommand\arraystretch{1.1}
							\begin{tabular}{l|c|cc}
								\thickhline
								\rowcolor{mygray}
								Latent Synchronization &  \#Params& Sintel clean & Sintel final\\

								\hline\hline
                                None&11.27M&0.51&0.74\\
                                Vanilla FC Layer & 11.64M&0.50&0.73\\
                                FC w/ Similarity~\cite{ma2023image}&11.76M&0.49&0.71\\
								\hline
        \arrayrulecolor{gray}\arrayrulecolor{black}
        \textbf{Ours (Eq.~\ref{eq:latent})} & 11.90M &0.48&0.69\\
        \hline
						\end{tabular}}
						\setlength{\abovecaptionskip}{0.3cm}
						\setlength{\belowcaptionskip}{-0.1cm}
						\caption{\small{\textit{Latent Synchronization}}}
						\label{table:dispatch}
					\end{subtable}
 \vspace{-18pt}
\end{table*}
\!\!\!\textbf{Key Components Analysis.} We study the major components of ProtoFormer: \textit{Cross-Attention Prototyping} (\S\ref{subsec:cross-attention}) and \textit{Latent Synchronization} (\S\ref{subsec:latent-syn}). A \texttt{Base} model is designed without considering prototype updating and prototype-feature assignment. Shown in Table~\ref{table:key}, \texttt{Base} reaches 0.55 and 0.81 in average EPE. After adding \textit{Cross-Attention Prototyping}, substantial improvements are observed (\ie, 0.55 $\rightarrow$ 0.51 in clean pass), suggesting the efficacy of prototyping updating even without explicit prototype-feature assignment. Incorporating \textit{Latent Synchronization} into \texttt{Base} can observe a noticeable performance gain (\ie, 0.81 $\rightarrow$ 0.77 in final pass). Finally, the integration of the two techniques culminates in peak performance.
\\
\textbf{Cross-Attention Prototyping.} We then study the efficacy of \textit{Cross-Attention Prototyping} by comparing to different updating methods, including cosine similarity, conventional cross-attention~\cite{vaswani2017attention}, Criss cross-attention~\cite{huang2019ccnet} and $K$-Means~\cite{yu2022k}. From the efficient and effective perspectives, \textit{Cross-Attention Prototyping} outperforms competitive methods (see Table~\ref{table:recurrent}). We further study the iteration step $N$ in Table~\ref{table:number}, suggesting that the error progressively decreases from 0.52 to 0.48 when increasing $N$ from 1 to 4, and saturates at 4.
Considering the computation time, we set $N=3$ to strike the balance between performance and computational cost.
The number of prototypes $K$ plays a pivotal role in defining the central grouping points for motion features. We therefore investigate the variant of $K$ in Table~\ref{table:prototype-number}.
\\
\!\!\!\textbf{Latent Synchronization.} Next, we study our \textit{Latent Synchronization} in Table~\ref{table:dispatch}. With a standard setting without any prototype-feature corresponding (\ie, None), the model reports 0.74 in final pass. After applying a vanilla fully-connected layer to update the feature, the error decreases to 0.73.
Though inspiring, our proposed \textit{Latent Synchronization} with carefully anchored prototypes yields advanced performance across all ablative methods (\ie, 0.69).

\begin{figure}[t]
    \centering
\includegraphics[width=0.44\textwidth, height=2.2cm]{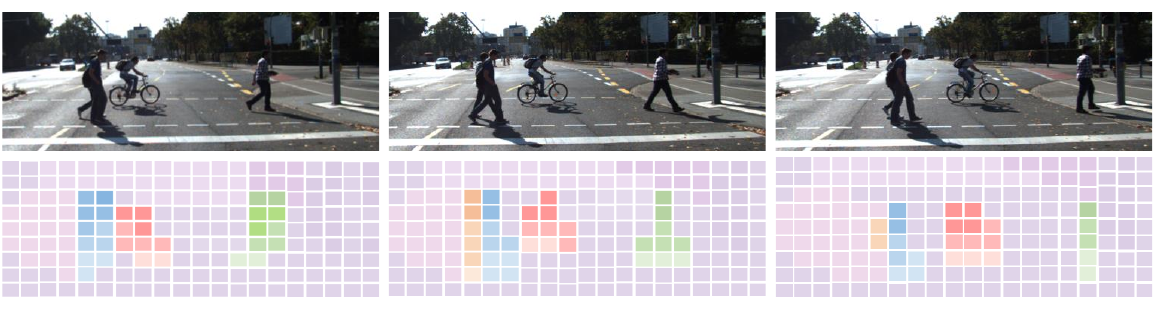}
    \vspace{-4mm}
    \caption{\small{
    \textbf{Visualization of proto-feature mapping,}
    which demonstrates  distinct prototypes with similar representations.}}
    \label{fig:visualize}
    \vspace{-9mm}
\end{figure}

\vspace{-1mm}
\textbf{Systemic Explainability.}
Finally, we investigate the prototype-feature corresponding map on optical flow in Fig.~\ref{fig:visualize}.
The systemic explainability hinges on the \textbf{Prototypes}, which emerge through the integration of probability density estimation within our cross-attention prototyping layer. The recursively optimized prototypes encapsulate the most characteristic features of the motion patterns within their respective density centers.
Through the visualization of feature correspondence estimation derived from the updated prototypes, we enhance the interpretability of the network and transparency of the model's decision-making process.

\vspace{-10pt}
\section{Conclusion}\label{sec:conclusion}
\vspace{-5pt}

We propose Prototypical TransFormer (ProtoFormer), a unified solution for motion tasks.
The motivation of integrating Transformer and prototype learning
leads us to innovate conventional self-attention, and propose \textit{Cross-Attention Prototyping}. Our \textit{Latent Synchronization} further refines the feature representations via prototype-feature association. Comprehensive empirical results show that ProtoFormer enjoys elegant architectural design, superior performance and systemic explainability. As a whole, we conclude that the findings from this work impart essential understandings and necessitate further exploration within this realm.

\section*{Impact Statement}\label{sec:social-impacts}

\textbf{Ethical Aspects.} We provide asset licenses and consents for the datasets we applied in our paper in supplementary material. All the datasets are publicly available for academic usage. Since our work does not involve data augmentation or the creation of new datasets, ethical concerns or biases within our proposed ProtoFormer are significantly minimized. We should also highlight that the introduction of prototypical learning offers a significant advantages in addressing photometric inconsistency, which in turn reduces the possible biases during training when lighting condition is restricted.

\textbf{Future Societal Consequences.} ProtoFormer introduces a universal understanding for motion tasks via prototypical learning, possessing strong performance gains over several state-of-the-art baselines. On positive side, our approach is valuable in various real-world applications (\eg, autonomous driving~\cite{cheng2023adversarial,prakash2021multi,cheng2024fusion, shao2023safety, cheng2022physical,cheng2023fusion}, robotics navigation~\cite{dev1997navigation, lookingbill2007reverse, song2022deep}), benefited from its transparency and efficiency.
Regarding potential negative social impacts, it is noteworthy that our ProtoFormer, akin to other discriminative classifiers, encounters challenges in addressing out-of-distribution/open-set problems~\cite{liang2022gmmseg}. Its utility in open-world scenarios should be further examined.

\section*{Acknowledgement}
We thank Prof.~Majid Rabbani for detailed feedback on drafts of the paper.
This research was supported by the National Science Foundation under Grant No. 2242243 and the DEVCOM Army Research Laboratory under Contract W911QX-21-D-0001. The views and conclusions contained herein are those of the authors and should not be interpreted as necessarily representing the official policies or endorsements, either expressed or implied, of the U.S. DEVCOM Army Research Laboratory (ARL), U.S. Naval Research Laboratory (NRL) or the U.S. Government.

\bibliography{example_paper}
\bibliographystyle{icml2024}

\clearpage
\appendix
\renewcommand{\thesection}{S\arabic{section}}
\renewcommand{\thetable}{S\arabic{table}}
\renewcommand{\thefigure}{S\arabic{figure}}
\setcounter{table}{0}
\setcounter{figure}{0}
\centerline{\textbf{SUMMARY OF THE APPENDIX}}

This supplementary contains additional experimental results and discussions of our ICML 2024 submission: \textit{Prototypical Transformer as Unified Motion Learners}, organized as follows:
\begin{itemize}
\item \S\ref{sec:config} provides detailed training configuration and testing configuration on optical flow and scene depth estimation.

\item \S\ref{sec:more_qualitative} provides more qualitative results and comparisons for optical flow and scene depth estimation.

\item \S\ref{sec:scene-depth-ablative} offers comprehensive ablation studies on scene depth estimation.

\item \S\ref{sec:proofs} provides detailed \textit{Proof} on the guarantees of $EM$ convergence.

\item \S\ref{sec:object-tracking} includes additional experiments on object tracking.
\item \S\ref{sec:Video-Stabilization} includes additional experiments on video stabilization.

\item \S\ref{sec:Reproducibility} discusses the
Pseudo codes.
\end{itemize}

\section{Training and Testing Configuration}\label{sec:config}
Our training methodology for ProtoFormer was adapted from established optical flow training protocols~\cite{jiang2021learning, huang2022flowformer}. Initially, the model underwent a pre-training phase on the FlyingChairs dataset~\cite{dosovitskiy2015flownet}, followed by an additional 120, 000 iterations on the FlyingThings dataset~\cite{mayer2016large}, a procedure we denote as ``C+T.''
Subsequently, the model underwent fine-tuning on a combined dataset encompassing FlyingThings~\cite{mayer2016large}, Sintel~\cite{butler2012naturalistic}, KITTI-2015~\cite{geiger2013vision}, and HD1K~\cite{kondermann2016hci}, referred to as ``C+T+S+K+H''. To optimize performance specifically for the KITTI-2015 benchmark~\cite{geiger2013vision}, we conducted a further fine-tuning phase on the KITTI-2015 dataset for 50, 000 iterations.
The training employed AdamW~\cite{loshchilov2017decoupled} optimizer and a one-cycle learning rate scheduler, with the peak learning rate set at $2.5 \times 10^{-4}$ for the FlyingChairs dataset and $1.25 \times 10^{-4}$ for the other datasets. Recognizing the sensitivity of transformer positional encodings to variations in image size, we adopted an image processing approach akin to that used in Perceiver IO~\cite{jaegle2021perceiver}. This involved cropping image pairs for flow estimation and subsequently tiling them to reconstruct complete flows. For depth prediction, we adhere to the architecture and configurations analogous to those employed for optical flow, as delineated in the respective headers. We initially adopt the VKITTI~\cite{cabon2020virtual} as a pretraining, and subsequently canonical Eigen split~\cite{eigen2014depth} and MPI Sintel dataset~\cite{butler2012naturalistic} to refine the model through fine-tuning, noted for its distinct edges and diverse motion intensities.
No additional data augmentation was used for the testing of all tasks.

\textbf{Loss Function.} For optical flow, A sequence loss is utilized for the training, which is defined over the sequence of flow predictions. For depth prediction, square root of the scale invariant logarithmic loss (SILog) is utilized for the training.

\vspace{-2mm}
\begin{table}[htb]
\resizebox{\columnwidth}{!}{%
\begin{tabular}{l|cc|cc}
\hline \thickhline
\rowcolor{mygray}
& \multicolumn{2}{c|}{LaSOT} & \multicolumn{2}{c}{TrackingNet} \\
\rowcolor{mygray}
\multirow{-2}{*}{Method} & Success $\uparrow$ & Precision $\uparrow$ & Success $\uparrow$ & Precision $\uparrow$ \\
\hline \hline
SiamFC~\cite{bertinetto2016fully} & 33.6 & 33.9 & 57.1 & 66.3 \\
MDNet~\cite{nam2016learning} & 39.7 & 37.3 & 60.6 & 56.6 \\
ECO~\cite{danelljan2017eco} & 32.4 & 30.1 & 55.4 & 49.2 \\
KYS~\cite{bhat2020know} & 55.4 & 55.8 & 74.0 & 68.8 \\
Ocean~\cite{zhang2020ocean} & 52.6 & 52.6 & 70.3 & 68.8 \\
TrDiMP~\cite{wang2021transformer} & 63.9 & 61.4 & 78.4 & 73.1 \\
TransT~\cite{chen2021transformer} & 64.9 & 69.0 & 81.4 & 80.3 \\
UniTrack~\cite{wang2021different} & 35.1 & 32.6 & 59.1 & 51.2 \\
UTT~\cite{ma2022unified} & 64.6 & 67.2 & 79.7 & 77.0 \\
UTT + \textcolor{gray}{\footnotesize{Ours}} & 64.8 & 67.4 & 80.0 & 77.2 \\
\hline
\end{tabular}
}
\vspace{-2mm}
\caption{Quantitative results on LaSOT and TrackingNet datasets. We are able to achieve competitive performance over state-of-the-art methods.}
\label{tab:object-tracking}
\end{table}

\section{More Qualitative Results
}\label{sec:more_qualitative}
\begin{figure*}
    \centering
    \includegraphics[width=0.91\textwidth, height=6.7cm]{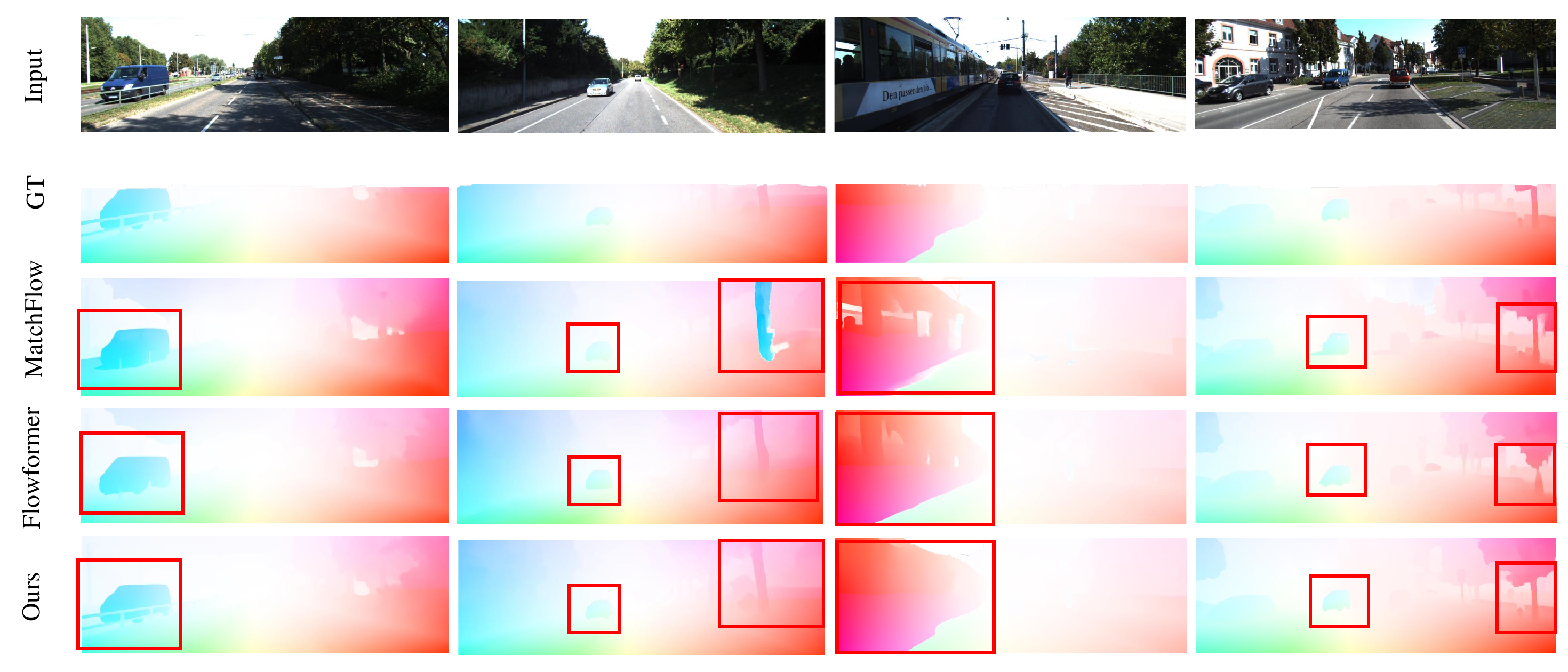}
    \caption{\textbf{More qualitative comparison on the KITTI test set.} The red boxes highlight the regions compared. Our method estimates more consistent and detailed flows.
}
    \label{fig:more_flow}
\end{figure*}

\begin{figure*}
    \centering
    \includegraphics[width=0.91\textwidth, height=7.2cm]{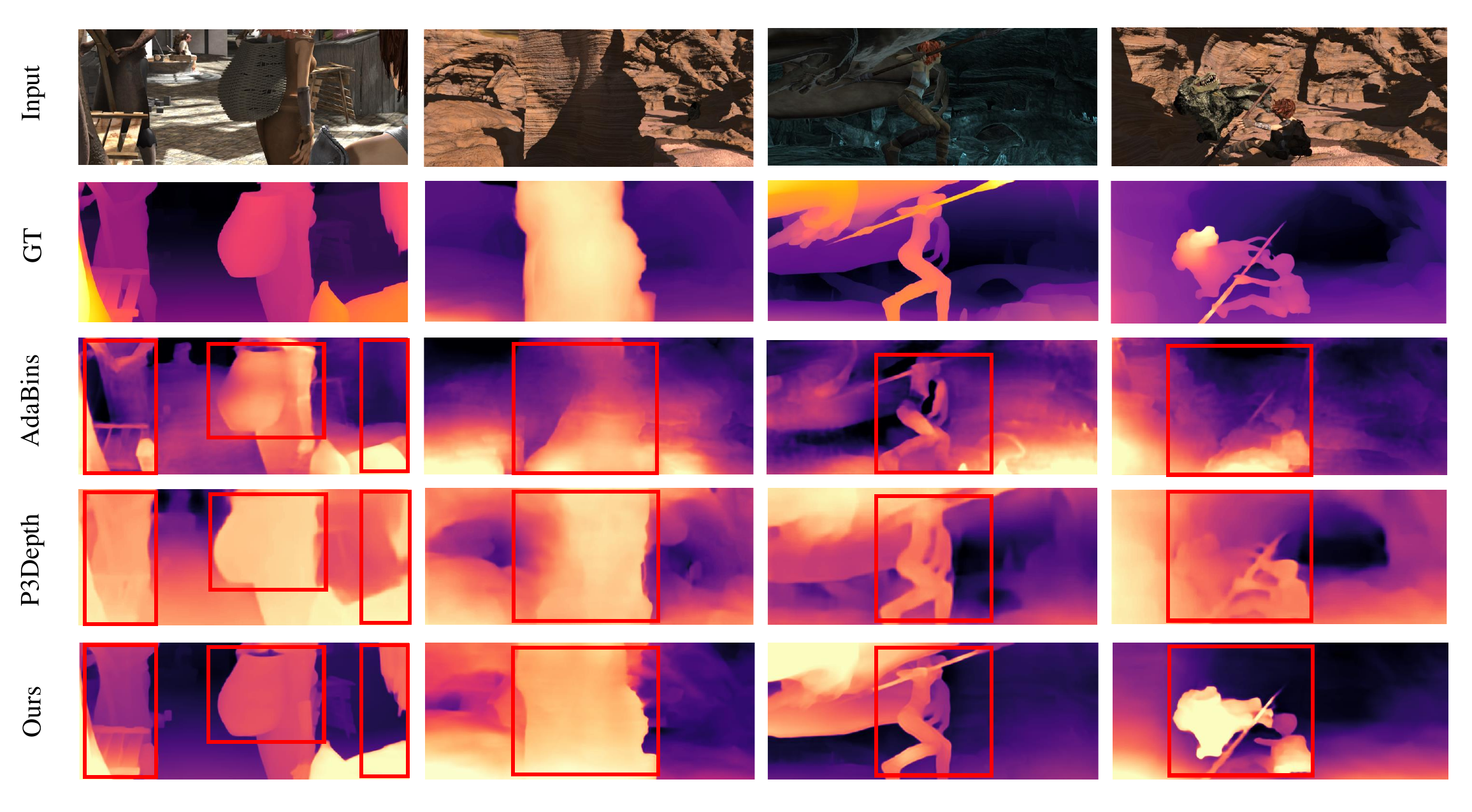}
    \caption{\textbf{More qualitative comparison on the Sintel test set.} The red boxes highlight the regions compared. Our method contributes to clearer depths without being affected by shadows or occlusions.
}
    \label{fig:more_depth}
\end{figure*}

We show more qualitative results on the main tasks, optical flow and scene depth, in Fig.~\ref{fig:more_flow} and Fig.~\ref{fig:more_depth}.

\begin{figure*}[t]
    \centering
    \includegraphics[width=0.92\textwidth, height=3.0cm]{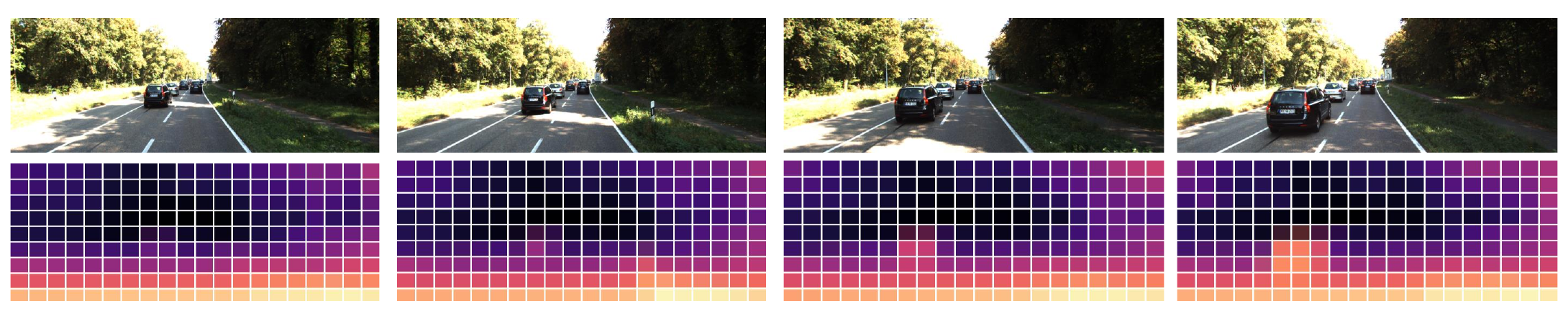}
    \vspace{-4mm}
    \caption{\small{
    \textbf{Visualization of proto-feature mapping in depth.} The map shows distinct prototypes with similar representations, indicating straightforward explainability.}}
    \label{fig:visualize_dep}
    \vspace{-4pt}
\end{figure*}

\begin{table*}[t]
    \caption{\small{\textbf{Ablative studies} on scene depth (see \S\ref{sec:scene-depth-ablative}).
    }}
     \vspace{-5pt}
     \begin{subtable}{0.45\linewidth}
						\captionsetup{width=.95\linewidth}
						\resizebox{\textwidth}{!}{
							\setlength\tabcolsep{4pt}
							\renewcommand\arraystretch{1.1}
							\begin{tabular}{l|c|cc}
								\thickhline
								\rowcolor{mygray}
								Algorithm Component &  \#Params& Abs Rel & RMSE\\

								\hline\hline
                                \texttt{Base}&9.63M & 0.074 & 2.835\\							\arrayrulecolor{gray}\hdashline\arrayrulecolor{black}	
                                ~~~$+$ Cross-Attention Prototyping &  11.57M & 0.067 & 2.742\\
                                ~~~$+$ Latent Synchronization & 10.26M & 0.071 & 2.819\\
								\hline
        \arrayrulecolor{gray}\arrayrulecolor{black}
        \textbf{ProtoFormer} (\textbf{All included})& 11.90M &0.062&2.716\\
        \hline
						\end{tabular}}
						\setlength{\abovecaptionskip}{0.3cm}
						\setlength{\belowcaptionskip}{-0.1cm}
						\caption{\small{Key Component Analysis}}
						\label{table:key2}
					\end{subtable}
          \begin{subtable}{0.46\linewidth}
						\captionsetup{width=.95\linewidth}
						\resizebox{\textwidth}{!}{
							\setlength\tabcolsep{4pt}
							\renewcommand\arraystretch{1.1}
							\begin{tabular}{l|c|cc}
								\thickhline
								\rowcolor{mygray}
								Variant Prototype Updating Strategy&  \#Params& Abs Rel & RMSE\\

								\hline\hline
                                Cosine Similarity&10.28M&0.066&2.775\\							
                                Vanilla Cross-Attention~\cite{vaswani2017attention}&14.88M& 0.066&2.767\\
                                Criss Cross-Attention~\cite{huang2019ccnet}&14.56M&0.065&2.750\\
                                $K$-Means~\cite{yu2022k}&11.81M&0.063&2.743\\
								\hline
        \arrayrulecolor{gray}\arrayrulecolor{black}
        \textbf{Cross-Attention Prototyping}& 11.90M &0.062&2.716\\
        \hline
						\end{tabular}}
						\setlength{\abovecaptionskip}{0.3cm}
						\setlength{\belowcaptionskip}{-0.1cm}
						\caption{\small{\textit{Cross-Attention Prototyping}}}
						\label{table:recurrent2}
					\end{subtable}
     \vspace{-0.5em}
     \begin{subtable}{0.30\linewidth}
     \vspace{-31pt}
						\captionsetup{width=.95\linewidth}
						\resizebox{\textwidth}{!}{
							\setlength\tabcolsep{4pt}
							\renewcommand\arraystretch{1.1}
							\begin{tabular}{c|c|cc}
								\thickhline
								\rowcolor{mygray}
								\#Iterations ($N$) &  \#Params& Abs Rel& RMSE\\

								\hline\hline
                                1&\multirow{4}{*}{11.90M}&0.070&2.801\\							
                                2&&0.065&2.737\\
                                \textbf{3}&  &0.062&2.716\\
                                4&&0.061&2.713\\
        \hline
						\end{tabular}}
						\setlength{\abovecaptionskip}{0.3cm}
						\setlength{\belowcaptionskip}{-0.1cm}
						\caption{\small{Number of Iterations}}
						\label{table:number2}
					\end{subtable}
     \begin{subtable}{0.30\linewidth}
     \vspace{-31pt}
						\captionsetup{width=.95\linewidth}
						\resizebox{\textwidth}{!}{
							\setlength\tabcolsep{4pt}
							\renewcommand\arraystretch{1.1}
							\begin{tabular}{c|c|cc}
								\thickhline
								\rowcolor{mygray}
								\#Prototypes ($K$) &  \#Params& Abs Rel & RMSE\\
								\hline\hline
                                10 & 8.95M & 0.070&2.775\\				
                                50 & 9.78M & 0.067 & 2.734\\
                                \textbf{100}& 11.90M &0.062&2.716\\
                                200 & 14.21M &0.064 & 2.720 \\
								\hline
						\end{tabular}}
						\setlength{\abovecaptionskip}{0.3cm}
						\setlength{\belowcaptionskip}{-0.1cm}
						\caption{\small{Number of Prototypes}}
						\label{table:prototype-number2}
					\end{subtable}
          \begin{subtable}{0.40\linewidth}
						\captionsetup{width=.95\linewidth}
						\resizebox{\textwidth}{!}{
							\setlength\tabcolsep{4pt}
							\renewcommand\arraystretch{1.1}
							\begin{tabular}{l|c|cc}
								\thickhline
								\rowcolor{mygray}
								Latent Synchronization &  \#Params& Abs Rel & RMSE\\

								\hline\hline
                                None&11.27M&0.067&2.732\\
                                Vanilla FC Layer & 11.64M&0.064&2.724\\
                                FC w/ Similarity~\cite{ma2023image}&11.76M&0.063&2.718\\
								\hline
        \arrayrulecolor{gray}\arrayrulecolor{black}
        \textbf{Ours}& 11.90M &0.062&2.716\\
        \hline
						\end{tabular}}
						\setlength{\abovecaptionskip}{0.3cm}
						\setlength{\belowcaptionskip}{-0.1cm}
						\caption{\small{\textit{Latent Synchronization}}}
						\label{table:dispatch2}
					\end{subtable}
 \vspace{-22pt}
\end{table*}

\section{Ablation Studies on Scene Depth Estimation}\label{sec:scene-depth-ablative}
In \S4.3, we ablate comprehensively under the optical flow setting. In this section, we further report ablation studies on scene depth estimation for completeness.

\textbf{Key Components Analysis.} We study the two major components of ProtoFormer: \textit{Cross-Attention Prototyping} (\S\ref{subsec:cross-attention}) and \textit{Latent Synchronization} (\S\ref{subsec:latent-syn}).
Same to our paper, the \texttt{Base} model is designed without considering prototype updating and prototype-feature assignment. In Table~\ref{table:key2}, \texttt{Base} reaches 0.074 in Abs Rel and 2.835 in RMSE. Adding \textit{Cross-Attention Prototyping} gets substantial improvements (\ie, 0.074 $\rightarrow$ 0.067 in Abs Rel).
Considering \textit{Latent Synchronization} brings a performance gain (\ie, 0.074 $\rightarrow$ 0.071). Finally,
the integration of these two techniques reaches peak performance as ProtoFormer, which is consistent to the tendency in our paper.

\textbf{Cross-Attention Prototyping.} We also study the efficacy of \textit{Cross-Attention Prototyping} design by comparing to different updating methods.
For efficient and effective perspectives, our \textit{Cross-Attention Prototyping} outperforms competitive methods (see Table~\ref{table:recurrent2}). We further study the iteration step $N$ in Table~\ref{table:number2}, suggesting that when increasing $N$ from 1 to 4, the error progressively decreases from 0.070 to 0.061, and almost saturates at 4.
Considering the computation time in iterations, we set $N=3$ to strike the optimal balance between performance and computation.
Consistent to our paper, we investigate the variant of $K$ (\ie, number of prototypes) in Table~\ref{table:prototype-number2}.
We select the preferred setting at $K=100$.

\textbf{Latent Synchronization.} We further study our \textit{Latent Synchronization} in Table~\ref{table:dispatch2}. With a standard setting without any feature-prototype corresponding, the model achieves 0.067 in Abs Rel. Applying a vanilla fully-connected layer increases the performance to 0.064.
Though inspiring, our proposed \textit{Latent Synchronization} with carefully anchored prototypes yields advanced performance across all ablative methods (\ie, 0.062). \\

\begin{figure*}
    \centering
    \includegraphics[width=0.90\textwidth, height=16.0cm]{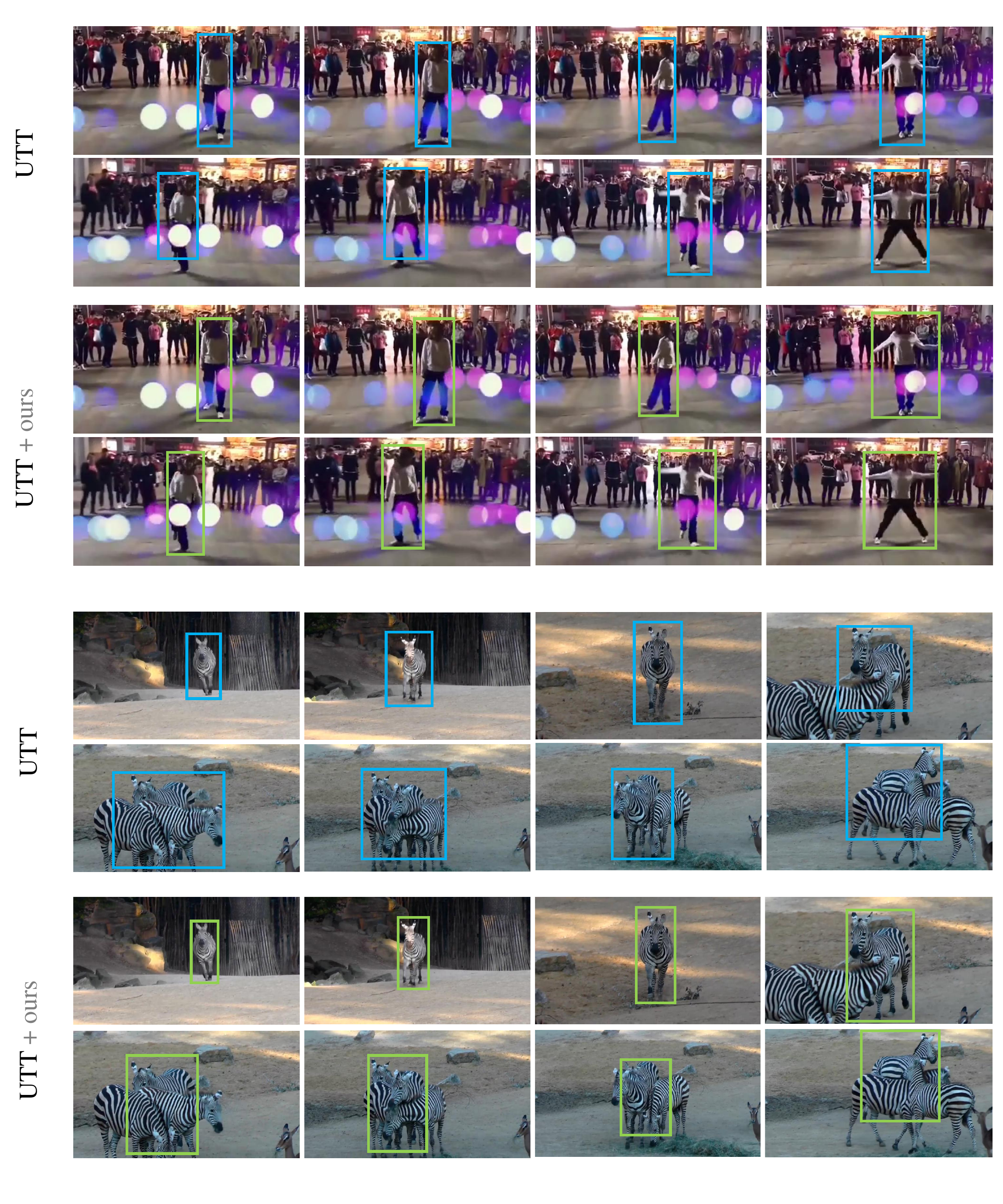}
    \caption{\textbf{Qualiative results on object tracking on LaSOT dataset.} With our enhanced method, the original ambiguous tracking due to illuminant changes and heavy occlusion becomes more accurate.
}
    \label{fig:tracking}
\end{figure*}

\begin{figure*}
    \centering
    \includegraphics[width=0.95\textwidth, height=14.0cm]{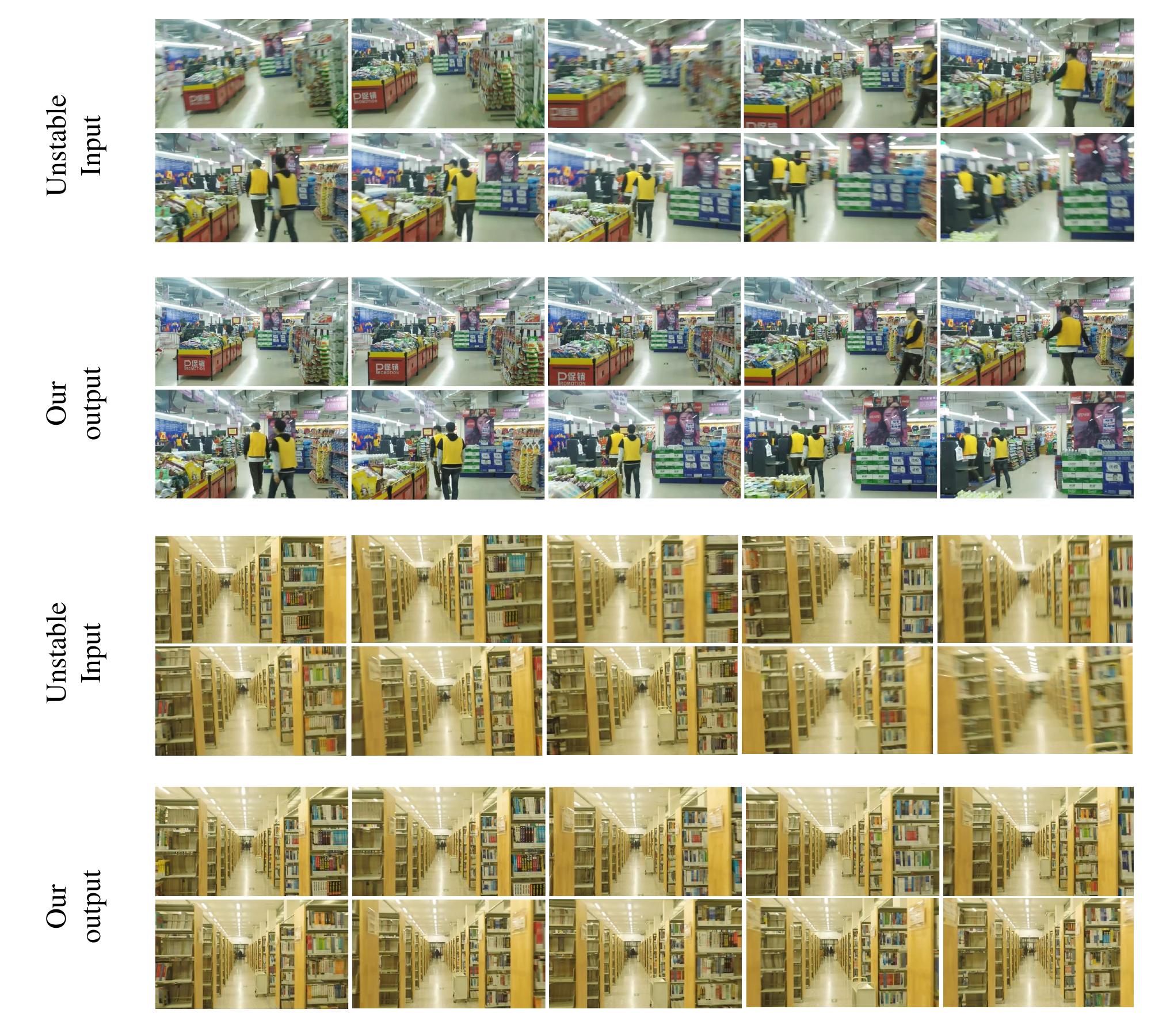}
    \caption{\textbf{Qualiative results on  video stabilization.} With our method enhanced, the original unstable video frames become much smoother and clearer.
}
    \label{fig:stable}
\end{figure*}

\section{Proof on the Guarantees of EM Convergence}\label{sec:proofs}

We first introduce the regularity condition, including a Euclidean ball of radius $r$ around the fixed point $\hat{\theta}$, set as:
\begin{equation}
\small
    \mathcal{B}_{2}(r;\theta):=\left\{\theta \in \Omega ~~\vert~~ ||\theta - \hat{\theta}||_{2} \leq r \right\}.
    \label{eq:ball}
\end{equation}
For simplicity, we define $\hat{\theta} = \argmax_\theta U(\cdot|\theta)$ while we have introduced our \textit{unified motion solution} in Eq.~\ref{eq:unified}.

For First-order Stability (FOS), the functions $U(\cdot|\theta)$ satisfy condition FOS($\gamma$) over $\mathcal{B}_{2}(r;\theta)$ if:
    \begin{equation}
    \begin{aligned}
        \small
    || \nabla U(M(\cdot|\hat{\theta})) - \nabla U(M(\cdot|\theta)) ||_{2} \leq \gamma ||\theta - \hat{\theta}||_{2},
    \label{eq:29}
    \end{aligned}
\end{equation}
for all $\theta \in \mathcal{B}_{2}(r;\hat{\theta})$. With radius $\gamma=0$, the condition of Eq.~\ref{eq:29} is always held at the fixed point $\hat{\theta}$. Extend further, by allowing for an always positive $\gamma$, given fixed point $\hat{\theta}$, Eq.~\ref{eq:29} would always hold in a local neighborhood $\mathcal{B}_{2}(r;\hat{\theta})$.

Under these conditions, we further guarantee the $EM$ operator to be locally contraction.

\begin{customthmFormal}{1}
    For $\gamma > 0$, and having $ 0 \leq \gamma \leq \lambda$, suppose the function $U(\cdot|\hat{\theta})$ is $\lambda$-strongly concave and FOS($\gamma$) holds for $\mathcal{B}_{2}(r;\hat{\theta})$, we have the $EM$ operator $M$ is contractive over $\mathcal{B}_{2}(r;\hat{\theta})$ as:
        \begin{equation}
    \begin{aligned}
        \small
    || M(\theta) - \hat{\theta}||_{2} \leq \frac{\gamma}{\lambda}||\theta - \hat{\theta}||_{2}
    \label{eq:not30}
    \end{aligned}
\end{equation}
for all $\theta \in \mathcal{B}_{2}(r;\hat{\theta})$. Intuitively, we can conduct that for any initial point $\theta^{(0)} \in \mathcal{B}_{2}(r;\hat{\theta})$, $\left\{\theta^{(n)}\right\}_{n=0}^{\infty}$ exhibits linear convergence.
Formally, we have:
\begin{equation}
    \begin{aligned}
        \small
    || \theta^{(n)} - \hat{\theta}||_{2} \leq (\frac{\gamma}{\lambda})^{n}||\theta^{0} - \hat{\theta}||_{2},
    \label{eq:30}
    \end{aligned}
\end{equation}
for all $n \in \left\{1, 2, ..., N \right\}$. Here we define $n$ in a finite set for intuitive reference below.
\label{theorem:ball-space}
 \end{customthmFormal}
Acknowledging the preliminary conditions in \textbf{\textit{Theorem \ref{theorem:ball-space}}}, we further present the proof for \textbf{\textit{{Proposition~\ref{prop:bound}}}} below.

\begin{proof}
For any iteration $n \in \left\{1,2, ... , N\right\}$, we have:
\begin{equation}
\small
    ||M_{m/N}(\theta^{(n)}) - M(\theta^{(n)})||_{2} \leq \epsilon_{M}(\frac{m}{N}, \frac{\delta}{N}),
    \label{eq:bound}
\end{equation}
with probability at least $1 - \frac{\delta}{N}$. Consequently, by a union bound over $N$, Eq.~\ref{eq:bound} holds uniformly with probability at least $1 - \delta$. It suffices to show that
\begin{equation}
    \small
    ||\theta^{(n+1)} - \hat{\theta}||_{2} \leq \kappa ||\theta^{(n)} - \hat{\theta}||_{2} + \epsilon_{M}(\frac{m}{N}, \frac{\delta}{N}),
    \label{eq:41}
\end{equation}
for each iteration $n \in \left\{1,2, ... , N\right\}$. When Eq.~\ref{eq:41} holds, we can iteratively show that:
\begin{equation}
    \begin{aligned}
        \small
    & ||\theta^{(n)} - \hat{\theta}||_{2} \leq \kappa ||\theta^{(n-1)} - \hat{\theta}||_{2} + \epsilon_{M}(\frac{m}{N}, \frac{\delta}{N}) \\
    & \leq \kappa \left\{ \kappa ||\theta^{(n-2)} - \hat{\theta}||_{2} + \epsilon_{M}(\frac{m}{N}, \frac{\delta}{N}) \right\}  + \epsilon_{M}(\frac{m}{N}, \frac{\delta}{N}) \\
    & \leq \kappa^{(n)} ||\theta^{(0)} - \hat{\theta}||_{2} + \left\{ \sum_{n=0}^{n-1} \kappa^{n} \right\} \epsilon_{M}(\frac{m}{N}, \frac{\delta}{N}) \\
    & \leq \kappa^{n} ||\theta^{(0)} - \hat{\theta}||_{2} + \frac{1}{1-\kappa} \epsilon_{M}(\frac{m}{N}, \frac{\delta}{N}).
    \label{eq:42}
    \end{aligned}
\end{equation}
The final step follows by summing the geometric series.

Thus, we need to prove Eq.~\ref{eq:41} via induction on the iteration number. Start with $n=1$, we have:
\begin{equation}
    \begin{aligned}
        \small
     & ||\theta^{(1)} - \hat{\theta}||_{2} = ||M_{m/N}(\theta^{(0)}) - M(\theta^{(n)})||_{2} \\
     & \mathbf{(I)} \leq ||M(\theta^{(0)}) - \hat{\theta}||_{2} + ||M_{m/N}(\theta^{(0)}) - M(\theta^{(0)})||_{2} \\
     & \mathbf{(II)} \leq \kappa ||M(\theta^{(0)}) - \hat{\theta}||_{2} + \epsilon_{M}(\frac{m}{N}, \frac{\delta}{N}),
    \label{eq:43}
    \end{aligned}
\end{equation}
where step \textbf{(I)} follows by the triangle inequality, step \textbf{(II)} follows from Eq.~\ref{eq:41}, and the contractivity of the operator applied to $\theta^{(0)} \in \mathcal{B}_{2}(r;\hat{\theta})$. In the induction from $n \rightarrow n+1$, suppose that $||\theta^{(n)} - \hat{\theta}||_{2} \leq r$, and the bound holds (\ie, Eq.~\ref{eq:41}) for iteration $n$. The same augment then implies that the bound from Eq.~\ref{eq:41} also holds for iteration $n + 1$, and that $||\theta^{(n+1)} - \hat{\theta}||_{2} \leq r$, thus completing the proof.
\end{proof}

\begin{table}[htb]
\resizebox{\columnwidth}{!}{%
\begin{tabular}{l|c|c}
\hline \thickhline
\rowcolor{mygray}
Method & Distortion Value $\uparrow$ & Stability Score $\uparrow$ \\
\hline \hline
StabNet~\cite{wang2018deep} &  0.83 & 0.75\\
StabNet + \textcolor{gray}{\footnotesize{Ours}} & 0.85 \textcolor{gray}{\footnotesize{(0.02 $\uparrow$)}} & 0.80 \textcolor{gray}{\footnotesize{(0.05 $\uparrow$)}} \\
\hline
PWStableNet~\cite{zhao2020pwstablenet} & 0.79 & 0.80\\
PWStableNet + \textcolor{gray}{\footnotesize{Ours}} & 0.82 \textcolor{gray}{\footnotesize{(0.03 $\uparrow$)}} & 0.83 \textcolor{gray}{\footnotesize{(0.03 $\uparrow$)}} \\
\hline
\end{tabular}
}
\vspace{-2mm}
\caption{Quantitative results on DeepStab dataset. We are able to achieve competitive performance over current methods.}
\label{tab:Video-Stabilization}
\end{table}

\section{Experiments on
Object Tracking
}\label{sec:object-tracking}

To further support our proposed ProtoFormer as a general solution to various tasks, we extend our design to object tracking
following \cite{ma2022unified}.
Intuitively, we follow~\cite{ma2022unified} and integrate the encoder and decoder into one object transformer where we replace the self-attention into our proposed cross-attention prototyping.

We evaluate our method on the testing splits of LaSOT~\cite{fan2019lasot} and TrackingNet~\cite{muller2018trackingnet} following common practices~\cite{ma2022unified, chen2023seqtrack}. Specifically, LaSOT~\cite{fan2019lasot} includes $1,400$ sequences: $1,120$ for training and $280$ for testing, respectively.
TrackingNet~\cite{muller2018trackingnet} contains $30K$ sequences with $511$ sequences for testing.
Success and Precision metrics are applied for performance evaluation.
We follow the same training schedule as~\cite{ma2022unified} for fairness.
As seen in Table~\ref{tab:object-tracking}, our approach achieves competitive results to current methods (\eg, 0.2\% and 0.2\% higher than UTT on Success and Precision on LaSOT dataset, respectively). Qualitative results are shown in Fig~\ref{fig:tracking}.

	\begin{algorithm}[ht]
	\caption{Pseudo-code of \textit{Cross-attention Prototyping} in a PyTorch-like style.}
	\label{alg:proto}
	\definecolor{codeblue}{rgb}{0.25,0.5,0.5}
	\lstset{
		backgroundcolor=\color{white},
		basicstyle=\fontsize{7.8pt}{7.8pt}\ttfamily\selectfont,
		columns=fullflexible,
		breaklines=true,
		captionpos=b,
		escapeinside={(:}{:)},
		commentstyle=\fontsize{7.8pt}{7.8pt}\color{codeblue},
		keywordstyle=\fontsize{7.8pt}{7.8pt},
	}
	\begin{lstlisting}[language=python]
"""
feats: output feature embeddings from regular projection, shape: (batch_size, channels, height, width)
P_0: initial cluster centers by adaptive pooling from the features, shape: (batch_size, num_clusters, dimension)
P: cluster centers, shape: (batch_size, num_clusters, dimension)
N: iteration number for recursive prototyping layer
"""

# One-step cross-attention prototyping in Eq.5
(:\color{codedefine}{\textbf{def}}:) (:\color{codefunc}{\textbf{one\_prototyping\_layer}}:)(Q, K, V):

    # E-step
    output = (:\color{codefunc}{\textbf{torch.matmul}}:)(Q, K.(:\color{codepro}{\textbf{transpose}}:)(-2, -1))
    M = (:\color{codefunc}{\textbf{torch.nn.functional.softmax}}:)(output, (:\color{codedim}{\textbf{dim}}:) = -2)

    # M-step
    P = (:\color{codefunc}{\textbf{torch.matmul}}:)(M, V)

    (:\color{codedefine}{\textbf{return}}:) P

# Iteratively cross-attention prototyping layer
(:\color{codedefine}{\textbf{def}}:) (:\color{codefunc}{\textbf{Cross\_Attention\_Prototyping}}:)(feats, P_0, N):

    Q = (:\color{codefunc}{\textbf{nn.Linear}}:)(P_0)
    K = (:\color{codefunc}{\textbf{nn.Linear}}:)(feats)
    V = (:\color{codefunc}{\textbf{nn.Linear}}:)(feats)
    P = P_0 + (:\color{codefunc}{\textbf{one\_prototyping\_layer}}:)(Q, K, V)

    (:\color{codedefine}{\textbf{for}}:) _ (:\color{codedefine}{\textbf{in}}:) (:\color{codedefine}{range}:)(N - 1):
        Q = (:\color{codefunc}{\textbf{nn.Linear}}:)(P)
        P = P + (:\color{codefunc}{\textbf{one\_prototyping\_layer}}:)(Q, K, V)

    (:\color{codedefine}{\textbf{return}}:) P

	\end{lstlisting}
\end{algorithm}

	\begin{algorithm}[ht]
	\caption{Pseudo-code of \textit{Latent Synchronization} in a PyTorch-like style.}
	\label{alg:sync}
	\definecolor{codeblue}{rgb}{0.25,0.5,0.5}
	\lstset{
		backgroundcolor=\color{white},
		basicstyle=\fontsize{7.8pt}{7.8pt}\ttfamily\selectfont,
		columns=fullflexible,
		breaklines=true,
		captionpos=b,
		escapeinside={(:}{:)},
		commentstyle=\fontsize{7.8pt}{7.8pt}\color{codeblue},
		keywordstyle=\fontsize{7.8pt}{7.8pt},
	}
	\begin{lstlisting}[language=python]
"""
feats: output feature embeddings from regular projection, shape: (batch_size, channels, height, width)
P: prototypes, shape: (batch_size, num_prototypes, dimension)
"""

# Latent sychronization in Eq.6
(:\color{codedefine}{\textbf{def}}:) (:\color{codefunc}{\textbf{latent\_sychronization}}:)(feats, P):

    max_value, max_index = (:\color{codefunc}{\textbf{similarity}}:)(feats, P).(:\color{codepro}{\textbf{max}}:)((:\color{codedim}{\textbf{dim}}:) = 1, (:\color{codedim}{\textbf{keepdim}}:) = True)
    mask = (:\color{codefunc}{\textbf{torch.zeros\_like}}:)((:\color{codefunc}{\textbf{similarity}}:)(feats, P))
    mask.(:\color{codepro}{\textbf{scatter\_}}:)(1, max_index, 1.)

    Q = (:\color{codefunc}{\textbf{nn.Linear}}:)(feats)
    K = (:\color{codefunc}{\textbf{nn.Linear}}:)(P)
    V = (:\color{codefunc}{\textbf{nn.Linear}}:)(P)

    feats += (:\color{codefunc}{\textbf{FFN}}:)((:\color{codefunc}{\textbf{attention\_layer}}:)(Q, K, V, (:\color{codedim}{\textbf{attn\_mask}}:) = mask))

    (:\color{codedefine}{\textbf{return}}:) feats


	\end{lstlisting}
\end{algorithm}

\section{Experiments on Video Stabilization}\label{sec:Video-Stabilization}

We further evaluate our method on video-based downstream task $-$ video stabilization, following common training configurations from~\cite{wang2018deep, zhao2020pwstablenet, lu2023transflow}.
Specifically, DeepStab~\cite{wang2018deep} contains 61 pairs of synchronized videos with diverse camera movements.
Distortion Value and Stability Score are applied for performance evaluation. In Table~\ref{tab:Video-Stabilization}, we report the results comparing to competitive methods (\ie, StabNet~\cite{wang2018deep}, PWStableNet~\cite{zhao2020pwstablenet}).
Specifically, we follow TransFlow~\cite{lu2023transflow}, aggregating the learned features from TransFlow's encoder (\ie, replacing the origin attention with our proposed cross-attention prototyping and latent synchronization) and the original encoder together for the later regressor.
Significant performance boost can be observed in both Distortion Value and Stability Score. Qualitative results are shown in Fig~\ref{fig:stable}.

\section{Pseudo-codes}\label{sec:Reproducibility}

ProtoFormer is implemented in Pytorch~\cite{paszke2019pytorch}. Experiments are conducted on eight NVIDIA A100-40GB GPUs. We provide the pseudo codes of our proposed ProtoFormer \textit{Cross-Attention Prototyping} in Algorithm~\ref{alg:proto} and \textit{Latent Synchronization} in Algorithm~\ref{alg:sync}.

\end{document}